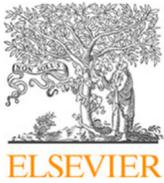
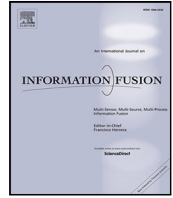

# A survey of large language models for healthcare: from data, technology, and applications to accountability and ethics

Kai He [a], Rui Mao [b], Qika Lin [a], Yucheng Ruan [a], Xiang Lan [a], Mengling Feng [a,*], Erik Cambria [b]

[a] *National University of Singapore, 119077, Singapore*
[b] *Nanyang Technological University, 639798, Singapore*

ABSTRACT

The utilization of large language models (LLMs) for Healthcare has generated both excitement and concern due to their ability to effectively respond to free-text queries with certain professional knowledge. This survey outlines the capabilities of the currently developed Healthcare LLMs and explicates their development process, to provide an overview of the development road map from traditional Pretrained Language Models (PLMs) to LLMs. Specifically, we first explore the potential of LLMs to enhance the efficiency and effectiveness of various Healthcare applications highlighting both the strengths and limitations. Secondly, we conduct a comparison between the previous PLMs and the latest LLMs, and summarize related Healthcare training data, learning methods, and usage. Finally, the unique concerns associated with deploying LLMs are investigated, particularly regarding fairness, accountability, transparency, and ethics. Besides, we support researchers by compiling a collection of open-source resources[1]. Summarily, we contend that a significant paradigm shift is underway, transitioning from PLMs to LLMs. This shift encompasses a move from discriminative AI approaches to generative AI approaches, as well as a move from model-centered methodologies to data-centered methodologies. **We determine that the biggest obstacle of using LLMs in Healthcare are fairness, accountability, transparency and ethics.**

## 1. Introduction

Recently, Large Language Models (LLMs) have emerged as a driving force in AI due to their impressive abilities in understanding, generating, and reasoning. The integration of LLMs into Healthcare represents a significant advancement in the application of AI toward improving clinical outcomes, conserving resources, and enhancing patient care. Healthcare researchers face persistent challenges such as diagnosing rare diseases, interpreting complex patient narratives, and planning personalized treatments. The advanced language processing capabilities of LLMs directly address these needs, offering more precise diagnostics and tailored treatment options. For example, Med-PaLM 2 [1] demonstrates expert-level accuracy on the US Medical Licensing Examination (USMLE). Besides, more general models such as GPT-4, GPT4-o and Llama series also demonstrate superior performance in a variety of healthcare-related tasks. These advancements expand LLM applications in healthcare while improving patient outcomes through greater accuracy and efficiency.

Initially, Pretrained Language Models (PLMs) include BERT [2] and RoBERTa [3] were developed for general NLP tasks and later adapted for healthcare applications. For simpler tasks, PLMs offer advantages over LLMs in terms of simplicity and efficiency when dealing with less complex cases. However, their use in healthcare was limited because they typically operated as single-task systems, lacking the capability to interact dynamically with complex medical data [4].

Then, the development of LLMs like GPT-3 represents a transformative evolution from PLMS to LLMs, as illustrated in Fig. 1. With over 100 billion parameters, GPT-3 demonstrates exceptional understanding and generating capabilities, which significantly enhance its functionality across various applications, including Healthcare [6]. These capabilities allow LLMs to process and analyze a broader array of data types, such as patient records, clinical notes, and research papers, to identify patterns and suggest potential diagnoses that might be overlooked by human clinicians [7]. Additionally, the integration of LLMs into Healthcare is further supported by their enhanced explainability and adaptability compared to PLMs. The introduction of Chain-of-Thought (CoT) processing in newer LLMs contributes to a more transparent AI decision-making process. This transparency is crucial in Healthcare settings, where understanding the rationale behind AI-generated decisions can foster greater trust and reliability among medical professionals in employing AI-powered tools [6].

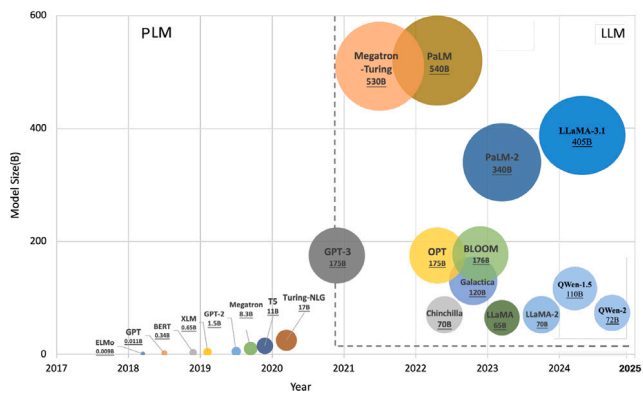

**Fig. 1.** The development road map from PLMs to LLMs. GPT-3 [5] marks a significant milestone in the transition from PLMs to LLMs, signaling the beginning of a new era both in general and Healthcare field.

Besides the aforementioned general abilities, many studies have tailored LLMs to address specific healthcare application tasks, marking a significant trend in this field. Understanding this trend is crucial for further advancing and diversifying healthcare applications. For instance, given that the healthcare field inherently involves multimodal data, some studies [8–10] have explored LLMs' capabilities to understand and analyze diverse medical images. Additionally, models like HuatuoGPT [11] demonstrate active inquiry capabilities, allowing for the extraction of more potential medical information. Other disease-specific LLMs, such as OphGLM [12] for ophthalmology and SoulChat [13] for mental health, highlight the versatility of LLMs in addressing targeted medical needs. Beyond these examples, the potential of LLMs in healthcare remains vast and largely untapped. Investing in the development of effective, ethical, and accountable LLMs is not only essential but also holds immense promise for practical and transformative benefits in healthcare.

This paper aims to inform readers about the latest developments in the field and offer comprehensive insights to those interested in using or developing healthcare LLMs. It covers various healthcare applications and provides a detailed summary of the underlying technology. We aims to provide insights about how different technologies affect different Healthcare-related tasks. Furthermore, as the capabilities of LLMs continue to improve, we contend that the challenges associated with applying AI in healthcare due to performance limitations are diminishing. Consequently, issues of fairness, accountability, transparency, and ethics are becoming more significant impediments to practical implementation. For this reason, we discuss these four critical issues in the context of employing LLMs and emphasize their importance.

Several surveys [7,14–16] have specifically examined the applications of large language models (LLMs) in medical and healthcare domains, emphasizing their potential benefits and limitations. However, these works lack in-depth technological analysis and fail to address critical issues such as accountability and ethics. Other surveys [17,18] include discussions on technological aspects but primarily focus on general LLM developments and evaluations, offering limited insights into their adaptation and application in healthcare settings. Some studies have a narrower focus. For instance, the study [19] concentrates solely on testing healthcare-specific LLMs, while [20] is limited to their applications in psychotherapy. Plus, former study [21] focused on Healthcare PLMs rather than LLMs. However, we provide a brief introduction to Healthcare PLMs as background information and then delve into the details of Healthcare LLMs. Our comprehensive analysis is anticipated to guide medical researchers in making informed choices in selecting LLMs suitable for their specific needs. The organizational framework of this paper is shown as Fig. 2. Generally, our contributions can be summarized as:

- We propose a comprehensive survey of LLMs in Healthcare, outlining a evolution road map from PLMs to LLMs, updating readers on the latest advancements in the field.
- We compiled a detailed list of publicly available data, training costs, and task performances for Healthcare LLMs, which is useful for developers and users of private Healthcare LLMs.
- We explore key non-technical aspects of LLMs in Healthcare, like fairness, accountability, transparency, and ethics, which are vital for advancing Healthcare AI applications.

## 2. What LLMs can do for healthcare? from fundamental tasks to advanced applications

Numerous endeavors have been made to apply PLMs or LLMs to Healthcare. In the early stages, the studies primarily focused on fundamental tasks, due to the challenges of accessing diverse medical datasets, the complexity of the medical domain, and limitations of the models' capabilities. Based LLMs, the concept of Artificial General Intelligence (AGI) for Healthcare has been proposed, which has led to more practical and advanced applications in various aspects of the Healthcare field, as shown in Fig. 3. In this sections, we analyze what LLMs can do for Healthcare in detail, and mainly compare the strengths and weaknesses of LLMs and PLMs on different tasks to highlight the development from PLMs to LLMs.

### 2.1. NER and RE for healthcare

The initial step toward unlocking valuable information in unstructured Healthcare text data mainly involves Named Entity Recognition (NER) and Relation Extraction (RE). These two tasks are main tasks to achieve Information Extraction (IE), which provide fundamental information for a range of other Healthcare applications, such as medical entity normalization and coreference [22], medical knowledge base and knowledge graph construction [23], and entity-enhanced dialogue [24]. For example, by employing NER and RE tasks, the Healthcare knowledge databases Drugbank[1] [25] and Unified Medical Language System (UMLS) are constructed, which facilitate various applications in Intellectual Healthcare.

In the early stages of research on NER with PLMs, a significant portion of studies focused on sequence labeling tasks. To accomplish this, PLMs-based approaches were employed to generate contextualized representations for individual tokens. In the case of RE tasks, the extracted entity pairs' representations were typically fed into a classifier to determine the existence of relations between the given entities. In the era of LLMs, NER and RE have been improved to work under more complex conditions and more convenient usages. One example is LLM-NERRE [26], which combines NER and RE to handle hierarchical information in scientific text. This approach has demonstrated the ability to effectively extract intricate scientific knowledge for tasks that require the use of LLMs. These tasks often involve complexities that cannot be effectively handled by typical PLMs. Meanwhile, LLMs can effectively perform medical NER and RE well even without further training. The study [27] employed InstructGPT [28] to perform zero-/few-shot IE from clinical text, despite not being trained specifically for the clinical domain. The results illustrated that InstructGPT can perform very well on biomedical evidence extraction, medication status extraction, and medication attribute extraction. This observation supports the notion that LLMs can be applied with flexibility and efficiency.

---
[1] Drugbank is a comprehensive online database that provides information on drugs and drug targets. The most recent version (5.0) includes 9591 drug entries, such as 2037 FDA-approved small molecule drugs, 241 FDA-approved biotech drugs, 96 nutraceuticals, and over 6000 experimental drugs.

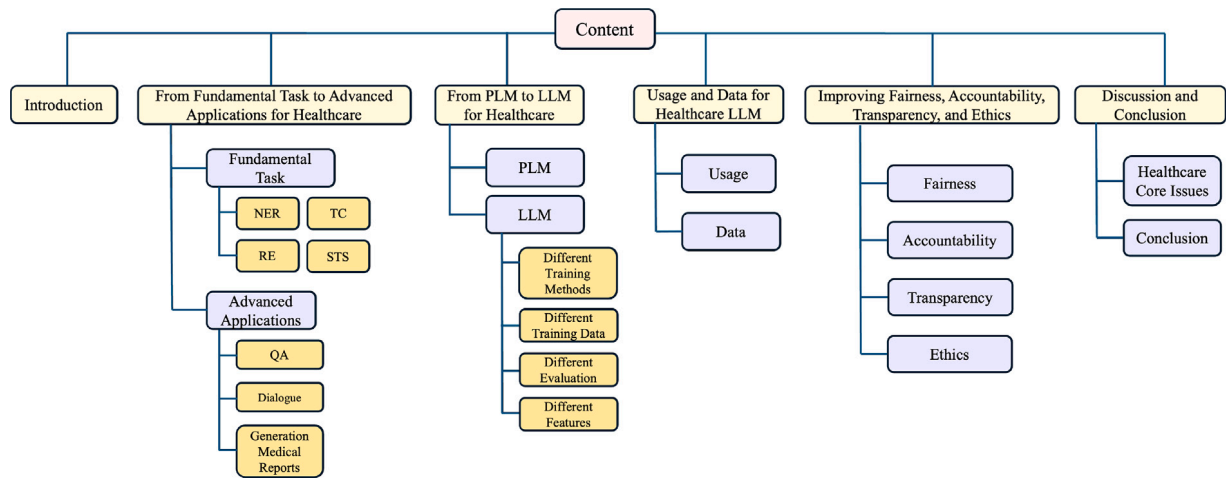

**Fig. 2.** The organizational framework for the content. Sections 3, 4 are technology details, while Sections 2, 5 are more valued for Healthcare professionals.

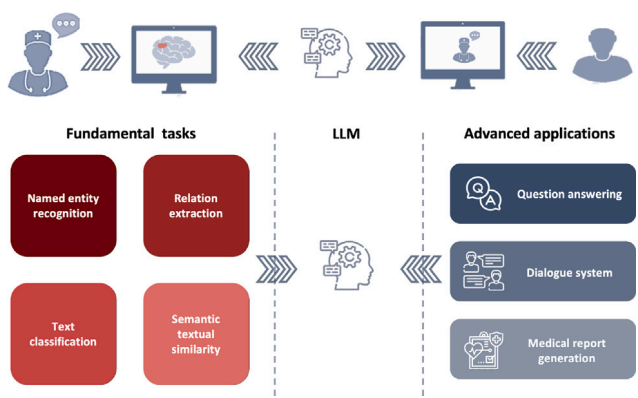

**Fig. 3.** LLMs for healthcare: from fundamental task to advanced applications.

Despite their capabilities, they still perform comparably to specially trained state-of-the-art (SOTA) PLMs, particularly in domains that involve professional terms and symbols. LLMs were trained on unlabeled data, with most of their knowledge derived from a vast amount of textual information. However, for domain-specific knowledge, such as specific types of named entities, LLMs' pragmatic understanding capabilities are likely to be less effective compared to PLMs that have been fine-tuned on labeled data. Overall, we argue that both PLMs and LLMs have distinct advantages in IE tasks.

### 2.2. Text classification for healthcare

Text Classification (TC) aims to assign labels to text with different lengths, such as phrases, sentences, paragraphs, or documents. In Healthcare research, a large amount of patient data is collected in the electronic format, including disease status, medication history, and treatment outcomes. However, these data can only be used with appropriate labels, while TC is one of the most commonly used technology. For example, a research study [29] proposed several methods, based on hybrid Long Short-Term Memory (LSTM) and bidirectional gated recurrent units(Bi-GRU) to achieve medical TC. The study [30] used TC to identify prescription medication mentioned in tweets and achieved good results by using PLMs. Also, some studies employ TC-based Sentiment Analysis (SA) to understand patient emotion or mental healthcare, aiming to provide more humanized treatments [31].

However, PLMs-based TC usually cannot satisfy explainable and reliable requirements in the Healthcare field, while LLMs-based TC mitigates these issues to some extent. For example, CARP [32] takes advantage of LLMs by introducing Clue And Reasoning Prompting to achieve better TC tasks. This study adopts a progressive reasoning strategy tailored to address the complex linguistic phenomena involved in TC. AMuLaP [33] is another example, which proposed Automatic Multi-Label Prompting for few-shot TC. By exploring automatic label selection, their method surpasses the GPT-3-style in-context learning method, showing significant improvements compared with previous PLMs-based results.

Unlike in general domains where LLMs and SOTA PLMs exhibit similar performance, LLMs demonstrate a clear advantage in Healthcare TC, which primarily due to the inherent complexity of special data. Healthcare texts are laden with specialized language, including technical terms, abbreviations, and jargon that are unique to the field. Moreover, the context in which these terms are used can significantly alter their meanings. For instance, the abbreviation "MI" might mean "mitral insufficiency" or "myocardial infarction", depending on the surrounding context. Given these conditions, Healthcare TC tasks require the integration of various types of data and an understanding of their interplay. This necessitates models that are not only summarize information but also reason contextually. LLMs are well-suited for these tasks due to their deeper contextual understanding and ability to handle complex interactions within the text, making them more effective for healthcare applications than PLMs.

### 2.3. Semantic textual similarity for healthcare

Semantic Textual Similarity (STS) is a way to measure how much two sentences mean the same thing or two documents are similar. In Healthcare, STS is often used to combine information from different sources, especially used for Electronic Health Records (EHR). The 2018 BioCreative/Open Health NLP (OHNLP) challenge [34] and the National NLP Clinical Challenges (n2c2) 2019 Track 1 show that STS can help reduce mistakes and disorganization in EHRs caused by copying and pasting or using templates. This means that STS can be used to check the quality of medical notes and make them more efficient for other NLP tasks. The study [35] proposed a new method using ClinicalBERT, which was a fine-tuned BERT-based method. The proposed iterative multitask learning technique helps the model learn from related datasets and select the best ones for fine-tuning. Besides, STS can be used for Healthcare information retrieval. For examples, if a patient ask question like "I was diagnosed with non-clear cell renal cell carcinoma, what are the chances of recurrence after cure? Give me evidence from relevant scientific literature", Our AI systems may need retrieval related database to find papers which contain similar semantic sentences. For doctor, when face patients who are difficult to diagnose, this technology can identify similar patients for doctors' reference.

When comparing PLMs and LLMs, we need to break down the situation to start some discussion. For short text semantic classification, SOTA PLMs and LLMs are comparable. This is primarily because such tasks contain less contextual information, meaning the advantages of LLMs in managing large context windows and understanding complex narrative structures are less pronounced. In such cases, the fundamental ability of both PLMs and LLMs to understand and interpret language at a basic level plays a more significant role, leading to similar levels of performance. On the other hand, for tasks like information retrieval, LLMs tend to be overly complex and resource-intensive for the role of a simple retriever. Typically, LLMs excel in directly generating responses or completing texts based on given inputs. In contrast, PLMs, which are generally more lightweight, are better suited for retrieving external knowledge. This distinction makes PLMs more practical for applications where quick, efficient retrieval of information is required without the additional overhead of generating new text content.

*2.4. Question answering for healthcare*

Traditionally, QA is a separate task that involves generating or retrieving answers for given questions. In Healthcare, QA can be very beneficial for medical professionals to find necessary information in clinical notes or literature, as well as providing basic Healthcare knowledge for patients. According to a report by the Pew Research Center [36], over one-third of American adults have searched online for medical conditions they may have. A strong QA system for Healthcare can significantly fulfill the consultation needs of patients. Many studies [21] explored how to adapt general PLMs to answer Healthcare questions, including designing special pertaining task, fine-tuning on Healthcare data, and introducing external Healthcare knowledge base. However, due to their limited language understanding and generation abilities [37], PLMs-based QA systems struggle to play a significant role in real-world Healthcare scenarios.

With the advent of powerful LLMs, prompt-based methods have been introduced to solve various tasks by formulating them as QA tasks, including NER, RE, and SA. Besides, LLMs have significantly improved typical QA tasks in Healthcare fields. For instance, Med-PaLM 2 [1] approached or exceeded state-of-the-art performance across MedMCQA [38], PubMedQA [39], and MMLU [40] clinical topics QA datasets. The study [41] investigated the use of ChatGPT, Google Bard, and Claude for patient-specific QA from clinical notes. Another study [42] proposed a retrieval-based medical QA system that uses LLMs in combination with knowledge graphs to address the challenge.

Visual Question Answering (VQA) has recently garnered significant attention in the Healthcare field for its potential to meet the diverse needs of both patients and healthcare professionals. By facilitating the interpretation of medical images through question answering, VQA holds great promise for aiding diagnostics and enhancing patient understanding through educational tools. One of the key challenges in this domain is the precise identification and comprehension of critical regions in medical images, such as masses, anomalies, and lesions. Equally vital is ensuring that the semantic representation of these regions aligns with the specific demands articulated in textual queries, enabling the generation of contextually relevant and medically accurate responses. For example, The study [43] introduces a novel multiple meta-model quantification method for medical VQA tasks. This method effectively learns meta-annotations and extracts meaningful features. It is designed to enhance metadata through auto-annotation, handle noisy labels, and generate meta-models that produce more robust and reliable features. Besides, MISS [44] presents an efficient multi-task self-supervised learning framework, which unifies the text and multimodal encoders to enhance the alignment of image-text features effectively. Moreover, MISS introduces a novel Transfer-and-Caption method, leveraging LLMs to expand the feature space of single-modal image datasets.

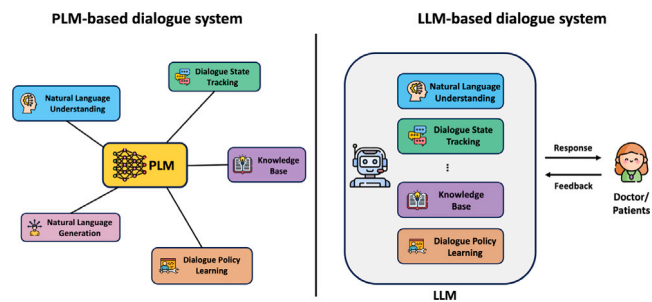

**Fig. 4.** The comparison between PLMs-based with LLMs-based dialogue system.

As one of their most outstanding ability, LLMs are obviously superior to PLMs on QA tasks. LLMs are increasingly being utilized to boost various real-world Healthcare applications, especially when considering only LLMs can support VQA tasks.

*2.5. Dialogue system for healthcare*

Chatbots have demonstrated promising potential to assist both patients and health professionals. The implementation of Healthcare Dialogue Systems can decrease the administrative workload of medical personnel and mitigate the negative consequences resulting from a shortage of physicians. Apart from the QA component, dialogue systems are generally classified into two categories: task-oriented and open-domain dialogue systems. The former is designed to address specific issues for Healthcare, such as hospital guides or medication consultations. In contrast, open-domain dialogue systems prioritize conversing with patients without any specific tasks. These systems are usually used as chatbots to provide emotional support, or mental health-related applications [45]. For example, the study [46] shows that patients who participated in a telehealth project had lower scores for depression, anxiety, and stress, and experienced 38% fewer hospital admissions. In the early stages, the study [47] proposed an ontology-based dialogue system that supports electronic referrals for breast cancer, which can handle the informative responses of users based on the medical domain ontology. Another study KR-DS [48] is an end-to-end knowledge-routed relational dialogue system that seamlessly incorporates a rich medical knowledge graph into topic transitions in dialogue management. One of the most notable feature is that PLMs-based dialogue systems often comprise multiple sub-modules, including dialogue management, nature language understanding, or knowledge introduction modules. Each individual sub-module within the overall system has the potential to become a bottleneck, thereby restricting the system's practical applications.

In the case of LLM-based dialogue systems, the original pipeline system can be transformed into an end-to-end system leveraging a powerful LLM [17], as shown in Fig. 4. By utilizing an LLM, the remaining task involves aligning the system with human preferences and fine-tuning it for specific fields, without the need for many extra sub-modules, and achieving some advanced abilities that PLMs can hardly do. For example, a new approach [49] was proposed to detect depression, which involves an interpretable and interactive system based on LLMs. The proposed system not only provides a diagnosis, but also offers diagnostic evidence that is grounded in established diagnostic criteria. Additionally, users can engage in dialogue with the system, which allows for a more personalized understanding of their mental state based on their social media content. Chatdoctor [50] is a specialized LLMs designed to overcome the limitations observed in the medical knowledge, which can utilize real-time information from online sources to engage in conversations with patients.

**Table 1**
Summarization about the strengths and weaknesses of PLMs and LLMs by different tasks.

| Task | PLMs features | LLMs features | Comparison |
| --- | --- | --- | --- |
| Information extraction | Need labeled data | Zero-/few-shot | Have their own unique strengths |
| Text classification | Easy to adapt | Explainable and reliable | LLMs have a slight advantage |
| Semantic textual similarity | Skilled at short contexts and fundamental tasks | Skilled at long contexts and complex tasks | Depend on text length |
| Question answering | Limited language understanding and generation abilities | Better inherent professional knowledge | LLMs have a significant advantage |
| Dialogue system | Consist of multiple components | End-to-end system | LLMs have a significant advantage |
| Report generation | Limited generation abilities and only single modality | Multimodal LLMs | LLMs have a significant advantage |

*2.6. Generation of medical reports from images*

Medical reports are of significant clinical value to related specialists, but the process of writing them can be tedious, time-consuming and error-prone for inexperienced ones. Therefore, the automatic generation of medical reports has emerged as a promising research direction in the field of Healthcare. This capability can assist specialists in clinical decision-making and reduce the burden of report writing by automatically drafting reports that describe both abnormalities and relevant normal findings. Additionally, related models are expected to assist clinicians by pairing text reports with interactive visualizations, such as highlighting the region described by each phrase.

In an early stage, the study [51] proposed a data-driven method that combines a CNN to predict medical tags and generate a single sentence report. However, a single-sentence report is limited to real medical scenes. To generate multi-sentence reports, the study [52] proposed a multi-level recurrent generation model, which fused multiple image modalities by focusing on the front and later views. Most recently proposed models for automated report generation rely on multimodal technology implemented by LLMs, which can support more advanced applications. For example, VisualGPT [53] utilizes linguistic knowledge from LLMs and adapts it to new domains of image captioning in an efficient manner, even with small amounts of multimodal data. ChatCAD [54] introduced LLMs into medical-image Computer Aided Diagnosis (CAD) networks. Their proposed framework leverages the capabilities of LLMs to enhance the output of multiple CAD networks, including diagnosis networks, lesion segmentation networks, and report generation networks. Their results show that ChatCAD achieved significant improvements under various measures compared with the other two report-generation methods (R2GenCMN [55] and CvT2DistilGPT2 [56]). ChatCAD+ [57] is a multimodal system that addresses the writing style mismatch between radiologists and LLMs. The system is designed to be universal and reliable, capable of handling medical images from diverse domains and providing trustworthy medical advice by leveraging up-to-date information from reputable medical websites. For such a complex task, LLMs clearly outperforms PLM by a wide margin.

*2.7. Summary*

Based on the information provided, we summarize the strengths and weaknesses of PLMs and LLMs by different tasks in Table 1 and conclude the following points. For simpler fundamental tasks, the distinct advantages of LLMs are less apparent. However, as the complexity of advanced tasks increases, particularly those involving complex data conditions, requiring advanced semantic understanding, and comprehensive generative capabilities, LLMs begin to demonstrate their strengths. Besides, LLMs play an integral role in specific sub-fields of Healthcare with enough further training, and turn to emphasis on the multimodal capability of LLMs, such as Healthcare data inherently consists of text, images, and time series data. By leveraging the strengths of LLMs, researchers and Healthcare professionals can harness the power of multiple modalities to improve diagnostic accuracy and patient care.

Beyond the accomplishments already discussed, several significant challenges remain for healthcare. A major obstacle is the complexity inherent in medical decision-making, which requires the incorporation of comprehensive patient information, including medical, psychological, and social aspects. While AI is proficient in analyzing data, it struggles with understanding complex human emotions and cultural nuances. This deficit is particularly evident in situations needing emotional support, such as during prolonged cancer care, where the empathetic engagement of healthcare professionals cannot be replicated by AI due to its inability to resonate emotionally.

Additionally, as AI becomes more embedded in healthcare, ethical and privacy issues intensify. Concerns about the handling of patient data, preserving privacy, and securing sensitive information are critical. Moreover, determining accountability in instances of diagnostic errors necessitates well-defined legal and ethical frameworks. Another concern is the unequal global distribution of technology, leading to a "digital divide". This divide risks leaving behind developing countries and economically disadvantaged areas, potentially worsening health disparities. AI also struggles with diseases characterized by ambiguous causes or intricate pathological processes. The effectiveness of AI is contingent on the extent of existing medical knowledge, and remains limited in fields that are not thoroughly understood. These challenges highlight the urgent need for collaborative efforts among professionals in healthcare, technology, law, and ethics globally to ensure that technological advancements are equitable, respectful of, and protective toward individual rights. Further discussion on these topics is available in Section 5.

**3. From PLMs to LLMs for healthcare**

Apart from the increasing model sizes, two significant developments from PLMs to LLMs are the transition from Discriminative AI to Generative AI and from model-centered to data-centered approaches. During the PLMs period, published PLMs were primarily evaluated on Natural Language Understanding (NLU) tasks, such as mentioned NER, RE, and TC. These studies are grouped as discriminative AI, which concentrates on classification or regression tasks instead of generation tasks. In contrast, generative AI generates new content, often requiring the model to understand existing data (e.g., textual instructions) before generating new content. The evaluation tasks of generative AI are usually QA and conversation tasks.

The second perspective is the change from model-centered to data-centered. Before the rise of LLMs, previous research focused on improving neural architecture to enhance the encoding abilities of proposed models. As neural models became increasingly larger, the over-parameterization strategy demonstrated promising abilities in learning potential patterns reserved in annotated datasets. Under such conditions, high-quality data played a more significant role in further enhancing various Healthcare applications. On the other hand, recent related developments present a multimodal trend, providing significant support to the data of EHRs, medical images, and medical sequence signals. Based on powerful LLMs, more existing and promising research and applications for Healthcare can be explored. Addressing the challenge of systematically collecting matched multimodal data holds significant importance. For such reason, we list detailed data usages and access links of each LLM in Section 3.2.

*3.1. PLMs for healthcare*

While our survey primarily concentrates on LLMs for Healthcare, it is important to acknowledge that previous studies on PLMs have played

**Table 2**

Summarization of training data and evaluation tasks for existing PLMs for Healthcare. The different training methods are delineated with a solid line and the training data are further delineated with a dashed line.

| Model name | Base | Para. (B) | Training data | Eval task | Date | Link |
| --- | --- | --- | --- | --- | --- | --- |
| BEHRT [58] | Transformer | – | CPRD, HES | Disease Prediction | 04/20 | Link |
| BioMegatron [59] | Megatron | 1.2 | PubMed | biomedical NER, RE, QA | 10/20 | Link |
| PubMedBERT [60] | BERT | 0.11 | PubMed | BLURB | 01/21 | Link |
| Bio-ELECTRA-small [61] | ELECTRA | 0.03 | PubMed | Biomedical NER | 03/20 | – |
| BioELECTRA [62] | ELECTRA | 0.03 | PubMed, PMC | BLURB, BLUE | 06/21 | Link |
| AraBERT [63] | BERT | 0.11 | Arabic Wikipedia, OSIAN | Arabic SA, NER, QA | 03/21 | Link |
| FS-/RAD-/GER-BERT [64] | BERT | 0.11 | Unstructured radiology reports | Chest Radiograph Reports Classification | 07/20 | Link |
| VPP [65] | BART | 0.14 | PubMed | Biomedical NER | 03/23 | Link |
| BioBART [66] | BART | 0.14 | PubMed | Biomedical EL, NER, QA, Dialogue, Summarization | 04/22 | Link |
| BioLinkBERT [67] | BERT | 0.34 | PubMed | BLURB, USMLE | 03/22 | Link |
| ELECTRAMed [68] | ELECTRA | 0.11 | PubMed | Biomedical NER, RE, and QA | 04/21 | Link |
| KeBioLM [69] | PubMedBERT | 0.11 | PubMed | BLURB | 04/21 | Link |
| BioFLAIR [70] | BERT | 0.34 | PubMed | Bio NER | 08/19 | Link |
| ouBioBERT [71] | BERT | 0.11 | PubMed, Wikipedia | BLUE | 02/21 | Link |
| SCIFIVE [72] | T5 | 0.77 | PubMed, PMC | Biomedical NER, RE, NIL, QA | 05/21 | Link |
| BioBERT [73] | BERT | 0.11 | PubMed, PMC | Biomedical NER, RE, QA | 05/19 | Link |
| BioALBERT-ner [74] | ALBERT | 0.18 | PubMed, PMC | Biomedical NER | 09/20 | Link |
| GreenCovidSQuADBERT [75] | BERT | 0.34 | PubMed, PMC, CORD19 | NER, QA | 04/20 | Link |
| Bio-LM [76] | RoBERTa | 0.34 | PubMed, PMC, MIMIC-III | 18 Biomedical NLP Tasks | 11/20 | Link |
| BioALBERT [77] | ALBERT | 0.03 | PubMed, PMC, MIMIC-III | 6 BioNLP Tasks | 04/22 | Link |
| BlueBert [78] | BERT | 0.34 | PubMed, MIMIC-III | BLUE | 06/19 | Link |
| ClinicalBert [79] | BERT | 0.11 | MIMIC-III | Hospital Readmission Prediction | 11/20 | Link |
| Clinical XLNet [80] | XLNet | 0.11 | MIMIC-III | PMV, Mortality | 11/20 | Link |
| MIMIC-BERT [81] | BERT | 0.34 | MIMIC-III | Biomedical NER | 08/19 | – |
| UmlsBERT [82] | BERT | 0.11 | MIMIC-III | MedNLI, i2b2 2006,2010, 2012, 2014 | 06/21 | Link |
| CharacterBERT [81] | BERT | 0.11 | MIMIC-III, OpenWebText, PMC | Medical NER, NLI, RE, SS | 10/20 | Link |
| Clinical KB-ALBERT [82] | ALBERT | 0.03 | MIMIC-III, UMLS | MedNLI, i2b2 2010, 2012 | 12/20 | Link |
| MedGPT [81] | GPT-2 | 1.5 | MIMIC-III, private EHRs | Disorder Prediction | 07/21 | – |
| KAD [83] | BERT | – | MIMIC-CXR | PadChest, ChestXray14, CheXpert and ChestX-Det10 | 03/23 | Link |
| Japanese-BERT [84] | BERT | 10.11 | Japanese EHR | Symptoms Classification | 07/20 | |
| MC-BERT [85] | BERT | 0.11 | Chinese EHR | Chinese Biomedical Evaluation benchmark | 08/20 | Link |
| BERT-EHR [86] | BERT | – | General EHR | Myocardial Infarction, Breast Cancer, Liver Cirrhosis | 03/21 | Link |
| Med-BERT [87] | BERT | 0.11 | General EHR | Disease prediction | 05/21 | Link |
| SAPBERT [88] | BERT | 0.11 | UMLS | MEL | 10/22 | Link |
| CODER [89] | mBERT | 0.34 | UMLS | MCSM, Medical RE | 02/22 | Link |
| AlphaBERT [90] | BERT | 0.11 | Discharge diagnoses | Extractive Summarization Task | 04/20 | Link |
| BioMed-RoBERTa [91] | RoBERTa | 0.11 | BIOMED | CHEMPROT, RCT | 05/20 | Link |
| RadBERT [92] | BERT | – | Radiology Report Corpus | Report Coding, Summarization | 05/20 | – |
| BioBERTpt [93] | BERT | 0.11 | Private clinical notes, WMT16 | SemClinBr | 11/20 | Link |
| RoBERTa-MIMIC [94] | RoBERTa | 0.11 | i2b2 2010, 2012, n2c2 2018 | i2b2 2010, 2012, N2C2 2018 | 12/20 | Link |
| CHMBERT [95] | BERT | 0.11 | Medical text data | Disease Prediction | 01/21 | – |
| Galén [96] | RoBERTa | 0.11 | Private clinical cases | CodiEsp-D, CodiEsp-P, Cantemist-Coding tasks | 05/21 | Link |
| Spanish-bert [97] | BERT | – | Spanish data | Spanish Clinical Case Corpus | 04/20 | – |
| French-BERT [98] | BERT | 0.11 | French clinical documents | DEFT challenge | 06/20 | – |
| ABioNER [99] | BERT | 0.11 | Arabic scientific literature | Arabic NER | 03/21 | – |
| SINA-BERT [100] | BERT | 0.11 | Online Persian source | Persian QA, SA | 04/21 | – |
| CT-BERT [101] | BERT | 0.11 | Tweet | COVID-19 Text Classification | 05/20 | Link |
| MentalBERT [45] | BERT | 0.11 | Reddit | Depression Stress, Suicide Detection | 10/21 | Link |

☆ PMV means prolonged mechanical ventilation prediction. NER means Named Entity Recognition, NLI means Natural Language Inference, RE means Relation Extraction, SS means Sentence Similarity. MCSM means medical conceptual similarity measure [102]. MEL means medical entity linking. EL means Entity Linking. For clarity, we only list parts of representative evaluation tasks. For the column of Para. (B), only the largest size is listed.

a foundational role in the development of LLMs. In this section, we sum up the key research points for Healthcare PLMs, namely (1) enhancing neural architectures, and (2) utilizing more efficient pre-training tasks. These two points will be compared with the distinct study focus of LLMs in Section 3.2, to further support the transition from discriminative AI to generative AI and from model-centered to data-centered.

• **Public Knowledge Bases.** There exist many Healthcare-related knowledge bases, such as UMLS [103], CMeKG [104], BioModels [105], and DrugBank [106]. Among them, UMLS is one of the most popular, which is a repository of biomedical vocabularies developed by the US National Library of Medicine. The UMLS has over 2 million names for 900,000 concepts from more than 60 families of biomedical vocabularies, as well as 12 million relations among these concepts. Based on this structured data, USMLE is organized and usually employed to test Healthcare LLMs. CMeKG [104] is a Chinese medical knowledge graph that has been constructed by referring to authoritative international medical standards and a wide range of sources, including clinical guidelines, industry standards, and medical textbooks. This knowledge graph serves as a comprehensive resource for medical information. Building upon the CMeKG, HuaTuo [107] utilizes diverse instructional data for its instruction tuning process.

• **Data for Instruction Fine-Tuning.** The aforementioned data typically consists of general text that is commonly used for pretraining PLMs or LLMs. However, when transitioning from PLMs to LLMs, instruction data becomes crucial to equip LLMs with the capability of following instructions effectively. Unlike PLMs, which primarily focus on next-word prediction, LLMs place greater emphasis on responding to specific instructions. By leveraging a sufficient amount of instruction data for fine-tuning, an LLM can appropriately generate the desired output. This emphasizes the importance of instruction-based training for LLMs to achieve accurate and contextually relevant responses.

For Healthcare PLMs, as shown in see Table 2, a majority of the models utilize the discriminative approach, predominantly built upon the BERT architecture. The rationale behind this architectural choice is

**Table 3**

Summarization of training data and evaluation tasks for existing LLMs for Healthcare. The different training methods are delineated with a solid line and the training data are further delineated with a dashed line. The color names represent popular evaluate datasets. More detail performance comparisons are shown in Table 4.

| Model name | Method | Training data | Evaluate datasets or tasks | Date | Link |
| --- | --- | --- | --- | --- | --- |
| GatorTron [108] | PT | Clinical notes | CNER, MRE, MQA | 06/22 | Link |
| GatorTronGPT [109] | PT | Clinical and general text | PubMedQA, USMLE, MedMCQA, DDI, BC5CDR | 05/23 | Link |
| Galactica [110] | PT+SFT | DNA, AA sequence | MedMCQA, PubMedQA, Medical Genetics | 11/22 | Link |
| Me LLaMA [111] | PT+SFT | PubMed, MIMIC-III, MIMIC-IV, MIMIC-CXR | MIBE benchmark [111] | 04/24 | Link |
| MedChatZH [112] | PT+SFT | Text Books, medical and general instructions | WebMedQA | 09/23 | Link |
| BioMistral [113] | PT+SFT | PubMed central data | MMLU, USMLE, MedMCQA, PubMedQA | 02/24 | Link |
| Visual Med-Alpaca [114] | PT+SFT | Medical QA | – | 04/23 | Link |
| Apollo [115] | PT+SFT | Books, clinical guidelines, encyclopedias. | XMedBench | 03/24 | Link |
| CancerLLM [116] | PT+SFT | Clinical notes, Pathology report | Cancer Diagnosis Generation, Cancer Phenotype Extraction | 06/24 | – |
| MedAlpaca [117] | SFT | Medical QA and dialogues | USMLE, Medical Meadow | 04/23 | Link |
| BenTsao [107] | SFT | Medical QA, Medical knowledge graph | Customed medical QA | 04/23 | Link |
| BianQue [118] | SFT | Medical QA | – | 04/23 | Link |
| Med-PaLM 2 [1] | SFT | Medical QA | MultiMedQA, Long-form QA | 05/23 | – |
| SoulChat [13] | SFT | Empathetic dialogue, Long text | – | 06/23 | Link |
| ChatDoctor [50] | SFT | Patient–doctor dialogues | iCliniq | 03/23 | Link |
| DoctorGLM [119] | SFT | Chinese medical dialogues | – | 04/23 | Link |
| OncoGPT [120] | SFT | Oncology conversations | Oncology Question Answering | 02/24 | Link |
| HuatuoGPT [11] | SFT | Conversation data and instruction | CmedQA, webmedQA, and Huatuo-26M | 05/23 | Link |
| Med-PaLM [121] | SFT | Medical data | MultiMedQA, HealthSearchQA | 12/22 | – |
| PMC-LLaMA [122] | SFT | Biomedical academic papers | PubMedQA, MedMCQA, USMLE | 04/23 | Link |
| HealAI [123] | SFT | Medical note data, instruction data | Medical Note Writing | 03/24 | – |
| BiMediX [124] | SFT | 1.3 million English-Arabic dataset | An Arabic-English benchmark | 02/24 | Link |
| Medical mT5 [125] | SFT | Multilingual medical corpus | Sequence Labeling, QA | 04/24 | Link |
| EpiSemoGPT [126] | SFT | Related publications | Predicting epileptogenic zones | 05/24 | – |
| MedAGI [10] | SFT | Public medical datasets and images | SkinGPT-4, XrayChat, PathologyChat | 06/23 | Link |
| Med-Flamingo [8] | SFT | Image-caption/tokens pairs | VQA-RAD, Path-VQA, Visual USMLE | 07/23 | Link |
| LLaVA-Med [9] | SFT | Multimodal biomedical instruction | VQA-RAD, SLAKE, PathVQA | 06/23 | Link |
| OphGLM [12] | SFT | Fundus image, knowledge graphs | Fundus diagnosis pipeline tasks [12] | 06/23 | Link |
| LLM-CXR [127] | SFT | MIMIC-CXR | Report generation, VQA, CXR generation | 05/23 | Link |
| JMLR [128] | SFT | MIMIC-IV dataset, medical textbooks, pubMed | USMLE, Amboss, MedMCQA, and MMLU-Medical | 02/24 | Link |
| ClinicalGPT [129] | SFT+RLHF | Medical dialogues and QA, EHR | MedDialog, MEDQA-MCMLE, MD-EHR, cMedQA2 | 06/23 | – |
| Polaris [130] | SFT+RLHF | Proprietary healthcare data | Healthcare conversational | 03/24 | – |
| Zhongjing [131] | PT+SFT+RLHF | Medical books, health records, clinical reports | CMtMedQA, Huatuo-26M | 08/23 | Link |
| Qilin-Med [132] | PT+SFT+DPO | Medical QA, plain texts, knowledge graphs | CMExam, CEval, Huatuo-26M | 04/24 | – |
| Aloe-Alpha [133] | PT+SFT+DPO | Medical QA, CoT, synthetic data | MultiMedQA, MedMCQA, USMLE, PubMedQA, etc. | 05/24 | – |

☆ ∗ means the study focuses on evaluating the Healthcare LLM, rather than proposing a new LLM. PT means pre-training, ICL means In-context-learning (no parameters updated), SFT means supervised fine-tuning, RLHF means reinforcement learning from human feedback, and DPO means Direct Preference Optimization.

evident: many typical Healthcare applications are classification tasks. These tasks range from NER in the biomedical domain to more specific challenges such as disease prediction and relation extraction. In addition, the methodology of fine-tuning (FT) stands out as the prevalent training methodology. This trend suggests a broader implication: while general pretrained models offer a foundational grasp of language, they require refinement through domain-specific data to excel in the applications of Healthcare. The choice of training datasets provides further support to the models' intent of achieving a holistic understanding of the medical domain.

Unlike PLMs, LLMs have the advantage of eliminating the need for FT and can directly infer at various downstream tasks. Moreover, the core research focus does not primarily revolve around improving neural architectures and developing more efficient pre-training tasks for Healthcare. Consequently, research on LLMs is garnering increased attention.

**Table 4**

The performance summarization for different Healthcare LLMs on three popular datasets.

| (%) | USMLE | MedMCQA | PubMedQA |
| --- | --- | --- | --- |
| FT BERT | 44.62 [67] | 43.03 [60] | 72.20 [67] |
| Galactica | 44.60 | 77.60 | 77.60 |
| PMC-LLaMA | 44.70 | 50.54 | 69.50 |
| GatorTronGPT | 42.90 | 45.10 | 77.60 |
| DoctorGLM | 67.60 | – | – |
| MedAlpaca | 60.20 | – | – |
| Codex | 60.20 | 62.70 | 78.20 |
| Med-PaLM | 67.60 | 57.60 | 79.00 |
| Med-PaLM | 67.60 | 57.60 | 79.00 |
| Aloe-Alpha | 71.01 | 64.47 | 80.20 |
| Med-PaLM 2 | 86.50 | 72.30 | **81.80** |
| GPT-4 | 86.70 | 73.66 | 80.40 |
| Human | **87.00** | **90.00** | 78.00 |

### 3.2. LLMs for healthcare

With the surge in general LLM studies, there has also been a notable development of LLMs specifically tailored for the Healthcare. In contrast to the emphasis on neural architecture designs and pretraining tasks in previous PLMs research, the studies on LLMs for Healthcare greater emphasis on collections of diverse, precise, and professional Healthcare data, and also data security and privacy protection. In the following sections, we present an overview and analysis of published Healthcare LLMs. For the sake of convenience, we have compiled the pertinent information in Tables 3 and 5. We categorize current LLMs based on their training methods, training data, evaluation, and distinct features, and offer detailed comparisons. Table 4 presents a summary of the performance for the three most popular datasets used to evaluate Healthcare LLMs, aimed at enabling more straightforward comparisons, and also offering a clear perspective on the current capabilities of excellent Healthcare LLMs.

• **Different Training Methods.** Unlike PLMs, the strategy of training LLMs from scratch is not popular for Healthcare LLMs. GatorTron [108] and GatorTronGPT [109] are only two Healthcare LLMs which training from scratch with only pretraining (PT). One of reason is that acquiring and properly anonymizing medical data for training involves navigating complex legal and ethical issues. Additionally, due to the specialized nature of medical data and the high demands for accuracy, training a model from scratch requires substantial computational resources and extremely large healthcare text, which will be more expensive than general LLMs. Compared with PLMs which require fewer parameters and less training data, the significance of PT method is in decline.

**Table 5**
Brief summarization of existing LLMs for Healthcare. Sorted in chronological order of publication.

| Model name | Size | Features |
| --- | --- | --- |
| GatorTron [108] | 8.9 | Training from scratch |
| Galactica [110] | 120 | Reasoning, Multidisciplinary |
| Med-PaLM [121] | 540 | CoT, Self-consistency |
| ChatDoctor [50] | 7 | Retrieve online, External knowledge |
| DoctorGLM [119] | 6 | Extra prompt designer |
| MedAlpaca [117] | 13 | Adapt to Medicine |
| BenTsao [107] | 7 | Knowledge graph |
| PMC-LLaMA [122] | 7 | Adapt to Medicine |
| Visual Med-Alpaca [114] | 7 | Multimodal generative model, Self-Instruct |
| BianQue [118] | 6 | Chain of Questioning |
| Med-PaLM 2 [1] | 340 | Ensemble refinement, CoT, Self-consistency |
| GatorTronGPT [109] | 20 | Training from scratch for medicine |
| LLM-CXR [127] | 3 | Multimodal, Chest X-rays |
| HuatuoGPT [11] | 7 | Reinforced learning from AI feedback |
| ClinicalGPT [129] | 7 | Multi-round dialogue consultations |
| MedAGI [10] | – | Multimodal |
| LLaVA-Med [9] | 13 | Multimodal, Self-instruct, Curriculum learning |
| OphGLM [12] | 6 | Multimodal, Ophthalmology LLM |
| SoulChat [13] | 6 | Mental Healthcare |
| Med-Flamingo [8] | 80 | Multimodal, Few-Shot medical VQA |
| Zhongjing [131] | 13 | Multi-turn Chinese medical dialogue |
| MedChatZH [112] | 7 | Traditional Chinese Medicine, Bilingual |
| JMLR [128] | 13 | RAG, LLM-Rank loss |
| BioMistral [113] | 7 | Multilingual, Model merging emphasis |
| BiMediX [124] | 47 | English and Arabic language |
| OncoGPT [120] | 7 | Real-world doctor-patient oncology dialogue |
| Polaris [130] | – | Several specialized support agents |
| HealAI [123] | 540 | RAG, Interactive Editing |
| Apollo [115] | 7 | Multilingual, Lightweight, Proxy tuning |
| Medical mT5 [125] | 3 | Multilingua |
| Qilin-Med [132] | 7 | Domain-specific pre-training, RAG |
| Me LLaMA [111] | 70 | Catastrophic Forgetting |
| EpiSemoGPT [126] | 7 | Predicting epileptogenic zones |
| Aloe-Alpha [133] | 8 | Synthetic CoT |
| CancerLLM [116] | 7 | Specifically for cancer |

Besides PT, the prevalent method for adapting a general LLM to a Healthcare LLM involves SFT. As shown in Table 3, 21 LLM studies only use SFT to tuning their models. In addition, Galactica, Me LLaMA, MedChatZH, BioMistral, Visual Med-Alpaca, and Apollo employ two-step training process, name PT first and then SFT. Among the above models, Galactica [110] is an early-stage study, which demonstrated effectiveness of SFT. This LLM is designed to handle the information overload in the scientific domain, including Healthcare. JMLR [128] introduces a method that enhances medical reasoning and question-answering by integrating SFT training method and information retrieval systems during the fine-tuning phase. This approach not only improves the model's ability to utilize medical knowledge effectively but also significantly cuts down on computational resources. Remarkably, JMLR required only 148 GPU hours for training. MedAlpaca [117] addresses privacy concerns in healthcare by employing an open-source policy for on-site implementation, which employs LoRA [148] for task-specific weight updates.

Further, the studies [129–132] use multiple advanced training technologies. Among them, Zhongjing [131] is a groundbreaking Chinese medical LLM that integrates PT, SFT, and RLHF to enhance the handling of multi-turn medical dialogues, particularly in Chinese medicine. Qilin-Med [132] is also a Chinese medical LLM enhanced through a multi-stage training methodology, including domain-specific PT, SFT, DPO, and Retrieval Augmented Generation (RAG).

• **Different Training Data.** Diverse and high-quality data is the one of core parts for Healthcare LLMs. In PLMs era, plain text dominates the training corpus for pretraining language models with the next word prediction task. When comes to Healthcare LLMs, QA pairs and dialogues one of more important data type, as shown in Line 12 to 20 in Table 3. This is due to the fact that the LLMs already have strong linguistic skills, as well as some degree of extra knowledge about the specifics of each domain. This attenuates the need to use specialized domain data to perform next word prediction tasks. More competitively, by using QA pairs and dialogues to construct instruction data, SFT can inject domain knowledge while enhancing the model's instruction compliance. Besides, some multimodal data (Line 27 to 30) and structured Electronic Health Record (EHR) database (Line 31 to 32) are also commonly used by SFT, which is other important training data. We can see a trend of synchronization between the different training methods and the training data. More details about training data can be seen in Section 4.2.

• **Different Evaluation.** Firstly, we investigate some work which focus in evaluate general LLMs for Healthcare tasks and categorize them into four folds: medical examination, medical question answering, medical generation, and medical comprehensive evaluation, which are summarized in Table 6. The medical examination form involves verifying model performance through standard medical tests or examinations. Differently, medical question answering involves utilizing questions posed or collected by human experts to make assessments. Medical generation focuses on generating new medical descriptions or knowledge based on a given input. The studies on medical comprehensive evaluation aim to provide assessments across various application scenarios rather than focusing on a single aspect. From conclusions of these studies, we can generally find that performance of specific tasks are satisfied, while more concerns are raised from non-technological parts, such as robustness, bias, and ethics. We further discussed these aspects in Section 5.

Secondly, we summarize evaluation parts from studies which propose Healthcare LLMs. For example, in Healthcare-related assessments, Galactica notably surpassed previous benchmarks with a 77.6% on PubMedQA and achieved 52.9% on MedMCQA. JMLR achieves 72.8% accuracy on the MMLU-Medical dataset and 65.5% on the MedMcQA dataset, surpassing the Meditron-70B and Llama2-13B with RAG, which scored 68.9% and 54.9% respectively.

Zhongjing [131] was evaluated using the CMtMedQA-test for multi-turn dialogues and the huatuo-26M for single-turn dialogues, focusing on three main dimensions—safety, professionalism, and fluency. Results show that Zhongjing excels in complex dialogue interactions, surpassing existing models like HuatuoGPT in these aspects by leveraging its diverse training approach. Qilin-Med achieved accuracies of 38.4% and 40.0% in the PT and SFT phases respectively on the CMExam test set. The integration of the RAG approach further enhanced its accuracy to 42.8% on CMExam. These advancements highlight Qilin-Med's capability in generating precise and contextually accurate responses, setting new benchmarks for medical LLMs, particularly in Chinese medical applications.

In summary, by integrating various training methods detailed in Table 3, we identify several overarching trends regarding the impact of different technologies on performance: (1) PT alone does not ensure high performance in LLMs; (2) SFT proves to be more crucial, with RLHF and DPO increasingly becoming important; (3) Techniques that reduce model size tend to result in some loss of performance.

• **Different Features.** Further, we discuss LLMs from features of **model sizes**, **language**, and **modality**. Model size is a crucial measure because it directly impacts the model's representation capabilities, generalization capacity, as well as the computational resources and training time required. We divide LLMs into three groups, extremely large (>70B), very large (13B-70B) and large (1B-12B). In this paper, there are 7/36 Healthcare LLMs are extremely large, 7/36 are very large, 19/36 are large. Med-PaLM [121] and HealAI [123] are two the largest Healthcare LLM with 540B parameters. Med-PaLM utilizes instruction prompt tuning for adapting LLMs to new domains with a few

**Table 6**
The Healthcare evaluation of LLMs.

| Categories | Studies | Models | Scenarios | #Num | Conclusions |
| --- | --- | --- | --- | --- | --- |
| Medical Ex. | [134] | ChatGPT | Primary Care | 674 | Average performance of ChatGPT is below the mean passing mark in the last 2 years. |
| | [135] | ChatGPT | Medical licensure | 220 | ChatGPT performs at the level of a third-year medical student. |
| | [136] | ChatGPT | Medical licensure | 376 | ChatGPT performs at or near the passing threshold. |
| Medical Q&A. | [137] | ChatGPT | Physician queries | 284 | ChatGPT generates largely accurate information to diverse medical queries. |
| | [138] | ChatGPT, GPT-4, Bard, BLOOMZ | Radiation oncology | 100 | Each LLM generally outperforms the non-expert humans, while only GPT-4 outperforms the medical physicists. |
| | [41] | ChatGPT, Claude | Patient-specific EHR | – | Both models are able to provide accurate, relevant, and comprehensive answers. |
| | [139] | ChatGPT | Bariatric surgery | 151 | ChatGPT usually provides accurate and reproducible responses to common questions related to bariatric surgery. |
| | [140] | ChatGPT | Genetics questions | 85 | ChatGPT does not perform significantly differently than human respondents. |
| | [141] | ChatGPT | Fertility counseling | 17 | ChatGPT could produce relevant, meaningful responses to fertility-related clinical queries. |
| | [142] | GPT-3.5, GPT-4 | General surgery | 280 | GPT-3.5 and, in particular, GPT-4 exhibit a remarkable ability to understand complex surgical clinical information. |
| | [143] | GPT-3.5, GPT-4 | Dementia diagnosis | 981 | GPT-3.5 and GPT-4 cannot outperform traditional AI tools in dementia diagnosis and prediction tasks. |
| Medical Gen. | [144] | ChatGPT | Gastroenterology | 20 | ChatGPT would generate relevant and clear research questions, but not original. |
| | [145] | ChatGPT, GPT-4 | Radiology report | 138 | ChatGPT performs well and GPT-4 can significantly improve the quality. |
| Medical Ce. | [146] | ChatGPT | Benchmark tasks | 34.4K | Zero-shot ChatGPT outperforms the state-of-the-art fine-tuned models in datasets that have smaller training sets. |
| | [147] | ChatGPT | Clinical and research | – | ChatGPT could potentially exhibit biases or be susceptible to misuse. |

☆ The Healthcare evaluation of LLMs includes Medical examination (Ex.), medical question answering (Q&A), medical generation (Gen.), and medical comprehensive evaluation (Ce.).

exemplars. This approach employs a shared soft prompt across multiple datasets, followed by a task-specific human-engineered prompt. Based on such extremely large size, Med-PaLM is evaluated on a 12-aspect benchmark and get satisfied results. For example, Med-PaLM and clinicians achieved a consensus of 92.6% and 92.9% respectively. Further, HealAI is based on Med-PaLM. However, there are no more details about its development. Med-PaLM 2 [1] is the second large Healthcare LLM with 340B parameters. Despite its smaller size compared to the original PaLM's 540B parameters, Med-PaLM 2 outperforms its predecessor [1]. Long-form answers from Med-PaLM 2 are evaluated for various quality criteria and often preferred over those from physicians and the original Med-PaLM model. Med-PaLM 2 also introduces ensemble refinement in its prompting strategy, enhancing answer accuracy by generating multiple reasoning paths to refine the final response. Besides Med-PaLM 2, Galactica and Me LLaMA [111] also have more than 100B parameters' models. It should notice that some smaller LLMs already outperform larger ones in general domains. This trend has not yet extended to Healthcare, but we anticipate that in the near future, smaller Healthcare LLMs will surpass the performance of older, larger models.

In the realm of language, English LLMs are predominantly mainstream. Following English, the second largest group of LLMs is designed for Chinese. BianQue, HuatuoGPT, BenTsao, SoulChat, DoctorGLM, MedChatZH, Zhongjing, and Qilin-Med are Chinese Healthcare LLMs. Among them, DoctorGLM is a pioneer Chinese LLM, focusing on cost-effective medical applications. DoctorGLM's training utilized the ChatDoctor dataset, translating medical dialogues using the ChatGPT API. Besides the above LLMs, there are also multilingual models, such as Apollo and Medical mT5.

Besides the above features, multimodal ability is another important development branch, as medical data inherently consists of diverse modalities such as patient medical records, radiographic images, and physiological signals. By integrating varied data types, multimodal models can enhance the understanding of complex medical conditions from multiple dimensions, enabling more accurate interpretations and diagnoses. For example, Visual Med-Alpaca [114] is a LLaMa-7B based open-source biomedical model that handles multimodal tasks by integrating medical "visual experts". It was trained using a collaboratively curated instruction set from GPT-3.5-Turbo and human experts, incorporating visual modules and instruction-tuning for tasks like radiological image interpretation and complex clinical inquiries. OphGLM [12] is a multimodal model tailored for ophthalmic applications, integrating visual capabilities alongside language processing. It was developed starting from fundus images, creating a pipeline for disease assessment, diagnosis, and lesion segmentation.

### 3.3. Summary

In this section, we present an overview of existing PLMs and LLMs in the Healthcare domain, highlighting their respective research focuses. Furthermore, we provide a comprehensive analysis of performance of Healthcare LLMs on benchmark datasets such as USMLE, MedMCQA, and PubMedQA as shown in Table 4. The intention behind this analysis is to showcase the progress in Healthcare QA development and offer a clear comparison between different Healthcare LLMs. In conclusion, two of the most robust LLMs identified in this analysis are Med-PaLM 2 and GPT-4. It is important to note that while GPT-4 is a general-purpose LLM, Med-PaLM 2 is specifically designed for Healthcare applications. Additionally, it is worth highlighting that the performance gap between LLM and human has significantly narrowed.

As mentioned earlier, one notable difference between PLMs and LLMs is that PLMs are typically discriminative AI models, while LLMs are generative AI models. Although there are auto-regressive PLMs like GPT-1 and GPT-2 also evaluated with classification tasks, auto-encoder PLMs have been more prominent during the PLMs period. As for LLMs, with their powerful capabilities, they have successfully unified various Healthcare tasks as QA or dialogue tasks in a generative way.

From a technological perspective, most PLM studies focus on improving neural architectures and designing more efficient pre-training tasks. On the other hand, LLM studies primarily emphasize data collection, recognizing the importance of data quality and diversity due to the over-parameterization strategy employed in LLM development. This aspect becomes even more crucial when LLMs undergo SFT to align with human desires. A study [1] reveals that the selection of mixed ratios of different training data significantly impacts the performance of LLMs. However, these mixed ratios of PT and SFT, often referred to as a "special recipe" from different strong LLM developers, are rarely publicized. Therefore, apart from SFT, we anticipate the emergence of more exciting and innovative methods for training LLMs, particularly those designed to handle unique features of Healthcare data.

Among the investigated Healthcare LLMs, most are derived from general LLMs. For these models, the SFT approach is the most commonly employed training technique. RLHF is less frequently utilized, with only MedAlpaca and HuatuoGPT adopting this method. The limited application of RLHF can be attributed to its high costs and stability challenges. RLHF relies on a reward model to guide training based on human feedback, but in the medical domain, obtaining expert input is significantly more expensive than in general fields. Additionally, inconsistent or noisy feedback can introduce reward variance, destabilizing the learning process. This issue is particularly pronounced in specialized areas like medicine, where expert opinions may diverge. Moreover, during RLHF, models risk catastrophic forgetting—losing previously learned information when new feedback contradicts prior knowledge. In medical applications, this can lead to the loss of critical information, compromising the model's reliability. Looking ahead, the development of more resource-efficient and stable RLHF algorithms is expected to enhance the performance and applicability of Healthcare LLMs.

Further, we have identified two emerging trends. Firstly, there is a growing exploration of multi-model approaches, including LLaVA-Med, MedAGI, OphGLM, Visual Med-Alpaca, and Med-Flamingo. Secondly, Chinese Healthcare LLMs are rapidly developing, with examples such as DoctorGLM, ClinicalGPT, SoulChat, BenTsao, BianQue, and HuatuoGPT. Finally, it is worth noting that many Healthcare LLM papers provide details about the prompts they used. This observation demonstrates the prompt brittleness, as different prompts can have a significant impact on the model's performance. Modifications in the prompt syntax, sometimes in ways that are not intuitive to humans, can lead to significant changes in the model's output. This instability is more matters for Healthcare than other general applications.

## 4. Usage and data for healthcare LLM

### 4.1. Usage

- **From Fine-tuning to In-context Learning.** In-context learning (ICL) offers promising benefits in healthcare by allowing LLMs to generate responses that mirror examples given by users. This method combines example demonstrations with test inputs to enhance the model's ability to utilize specific knowledge from these examples without needing to update parameters for specific healthcare data. ICL can be particularly effective in healthcare as it helps tailor these models to meet the precise requirements and expectations of medical professionals. Moreover, using examples can simplify interactions, as direct examples are often clearer and easier to understand than complex medical queries, which might not always capture the true intent of the user.

Nevertheless, the success of ICL in healthcare depends on various detailed factors like the similarity of inputs, the relevance of the labels, the format of the demonstrations, and how well the inputs and labels are paired. For example, it is vital that both the examples shown in training and the actual inputs used are from comparable medical situations. Also, the training labels must accurately reflect the labels used in real healthcare settings. The way the examples are presented must be carefully structured to ensure the model learns effectively from them. The study [149] investigates these aspects. While the precision of input-label mapping is less critical when label spaces are correctly aligned, inconsistencies in any of these areas can diminish the utility of ICL in real-world healthcare applications, as shown in Fig. 5. Therefore, meticulous attention to these parameters is essential to harness the full potential of ICL in enhancing diagnostic accuracy and efficiency in healthcare settings. However, Healthcare professionals are often not aware of these technology issues, resulting in LLMs not performing at their full potential.

- **From System 1 To System 2 – Chain-of-Thought.** According to the report [150], two distinct categories of Deep Learning systems exist, namely System 1 and System 2. System 1 encompasses the current applications of deep learning, including image recognition, machine translation, speech recognition, and autonomous driving. On the other hand, System 2 represents the future potential of deep learning, involving tasks such as reasoning, planning, and other logic-based and reasoning-oriented activities.

System-1 tasks in the field of NLP have been largely resolved, demonstrating significant progress. However, progress in System-2 tasks has been limited until recently when the emergence of advanced LLMs triggered a significant shift. The study [6] proposed the CoT prompting, which found it can significantly improve the reasoning and planning performance of LLM by adding a series of intermediate steps. Furthermore, the study [151] found that by just adding a sentence "Let's think step by step", the reasoning ability of LLMs can be significantly boosted. Later, there are many CoT studies [11,13,118] aiming to enhance the logical reasoning ability of LLM in various Healthcare applications.

The integration of CoT reasoning in Healthcare LLMs offers notable benefits for improving interpretability, particularly in complex decision-making processes such as clinical decision support systems. CoT enables models to break down decisions into explicit, step-by-step reasoning, making outputs more transparent and interpretable for healthcare professionals. However, these benefits come with trade-offs. The use of CoT can increase computational complexity and latency due to the need to generate detailed reasoning paths. In time-sensitive scenarios, such as healthcare emergencies, this added delay may limit the practicality of deploying CoT-enabled LLMs. To address this challenge, it is crucial to strike a balance by optimizing CoT reasoning to enhance transparency without sacrificing system responsiveness. Research on inference acceleration presents a promising approach to mitigating this issue, enabling faster processing while maintaining the interpretability advantages of CoT.

- **AI Agents.** The core idea behind recent AI agents is to build autonomous agent systems that utilize LLMs as their central controllers. These systems consist of several components, including Planning, Memory, Tool Use, and Action [152]. The planning component plays a crucial role in breaking down complex tasks into smaller and manageable sub-goals. This enables the agent to handle large tasks more efficiently by tackling them step by step. The Memory component provides the agent with the ability to store and retrieve information over extended periods. It typically utilizes an external vector store and fast retrieval mechanisms, allowing the agent to retain relevant knowledge and recall it as needed. With the Planning and Memory components in place, AI agents can take actions and interact with external tools. AutoGPT [153] is an example of such an autonomous agent system, which leverages GPT-4 to autonomously develop and manage operations. When provided with a topic, AutoGPT can think independently and generate steps to implement the given topic, along with implementation details. This shows the agent's autonomous ability to plan, utilize its memory, and take appropriate actions.

To our best knowledge, AI agents have not been widely adopted in the Healthcare field. However, we anticipate the development of more capable AI agent systems in this domain. For instance, it is possible to train specialized models for different medical processes, such as hospital guidance, auxiliary diagnosis, drug recommendation, and prognostic follow-up. These relatively small models can be integrated into a comprehensive AI medical system, where an LLM serves as the central controller. Additionally, specialized disease systems can be established for each department within the Healthcare system. The LLM can play a crucial role in determining which specialized disease systems should be involved in a particular case, resulting in effectively allocating resources and providing specialized care. Overall, the vision is to leverage AI agents and LLMs to create comprehensive and specialized AI systems in Healthcare, covering various medical processes and enabling efficient decision-making and patient care.

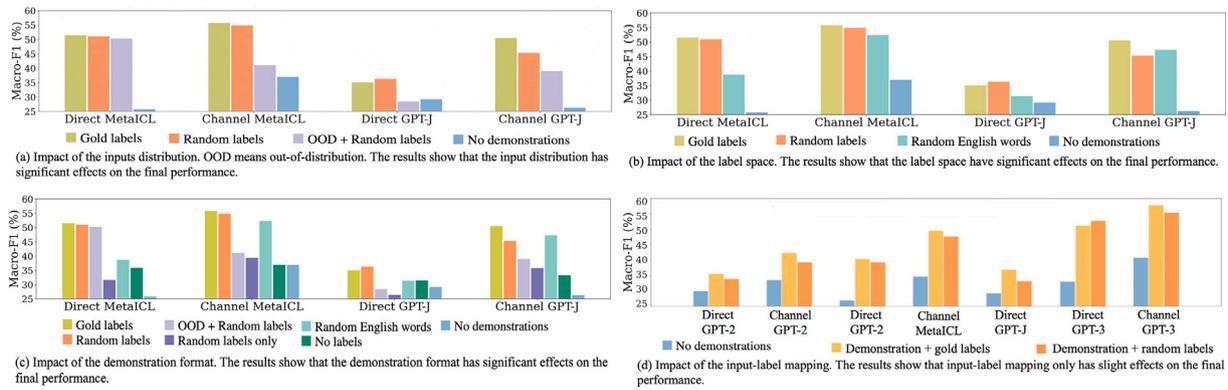

**Fig. 5.** What makes in-context learning work? ★ **The data of figures comes from the study** [149]. **We perform the proper arrangement and layout for discussions** ★. We only list the classification task (x-axis) here and sub-figure (d) shows parts of the original results for clarity.

*4.2. Healthcare training data*

As mentioned earlier, the transition from PLMs to LLMs brings a significant shift from a model-centered approach to a data-centered approach. Increasing the volume of pre-training data has become a key factor in enhancing the general capabilities of LLMs. In line with this, we have gathered and organized various datasets for training Healthcare LLMs, see shown in Table 7. Additional descriptions are listed below.

- **EHR.** The Medical Information Mart for Intensive Care III dataset (MIMIC III) is widely recognized as one of the most widely used EHR datasets. It encompasses a comprehensive collection of data from 58,976 unique hospital admissions involving 38,597 patients who were treated in the intensive care unit at the Beth Israel Deaconess Medical Center between 2001 and 2012. Furthermore, the dataset includes 2,083,180 de-identified notes that are associated with these admissions. MIMIC III provides valuable and extensive information, facilitating many PLMs and LLMs developments, such as MIMIC-BERT [81], GatorTron [108], and MedAGI [10].
- **Scientific Literature.** PubMed is a freely accessible search engine that provides access to the MEDLINE database, which contains references and abstracts related to life sciences and biomedical topics, with over 32 million citations for biomedical literature. The PubMed abstracts alone contain approximately 4.5 billion words, while the full-text articles available on PubMed Central (PMC) contribute around 13.5 billion words. These datasets consist of high-quality academic and professional text, making them particularly suitable for training Healthcare LLMs. Various PLM and LLMs, such as BioBERT [73], Bio-ELECTRA [182], GatorTron [108], and MedAlpaca [117], have been trained using PubMed data.
- **Web Data.** Web data includes any text we can obtain from the Internet. Social media is one of the most commonly used data types. Reddit is a popular online platform that combines social news aggregation, content rating, and discussion features. The platform is organized into user-created boards called "communities" or "sub-reddits", covering a broad range of topics. The study [183] crawled health-themed forums on Reddit to form COMETA corpus as LLMs training data. Tweets are also usually employed to collect data, and COVID-twitter-BERT [101], Twitter BERT [184], and TwHIN-BERT [185] are trained with these data.

In general, the most common sources of data for Healthcare LLMs include EHR, scientific literature, web data, and public knowledge bases. When considering the data structure, QA and dialogue are the most frequently encountered. Additionally, it is crucial to acknowledge the significance of multimodal data. Given that Healthcare domain inherently involves text, images, and time series data, multimodal LLMs offer a promising direction for further research.

Besides, we have summarized the relevant computation costs from existing studies in Table 8, which aims to provide clear assessment of computation requirements.

*4.3. Summary*

In this section, we first summarize usage for Healthcare LLMs, including ICL, CoT, and Agents. These technologies can further boost powerful capability of Healthcare LLMs without any expensive training process. Such non-parametric methods are also promising directions for further explorations to construct complete Healthcare AI systems. Also, we present a comprehensive overview about the data used for training LLMs, the volume often surpasses the capacity of human teams to manually perform quality checks. Consequently, data collection processes heavily rely on heuristic rules for selecting data sources and applying filters. In the context of LLM training, there are various data challenges to address, including the high cost of Healthcare data, contamination in benchmark data, personally identifiable information, and the mixture of domains during pre-training and fine-tuning tasks.

**5. Improving fairness, accountability, transparency, and ethics**

Fairness, accountability, transparency, and ethics are four important concerns in the AI domain. According to the study [186], *Fairness* holds paramount significance in guaranteeing that AI does not perpetuate or exacerbate established societal disparities; *Accountability* plays an important role in ensuring that individuals responsible for the conception and execution of AI can be held answerable for their decisions; *Transparency* assumes a critical role in ensuring that AI remains open to scrutiny and amenable to audits for possible biases or inaccuracies; *Ethics*, similarly, assumes a pivotal role in guaranteeing that AI is constructed and utilized in manners that align with prevailing social values and norms.

In the Healthcare domain, we believe that these four aspects are even more critical because the primary focus is on patient well-being and safety. In this context, the utmost importance lies in ensuring patients receive optimal care marked by equitable access to medical services. Additionally, the transparent and trustworthy nature of Healthcare decisions, the accountability in delivering accurate medical diagnoses and treatments, the safeguarding of patient confidentiality, and the adherence to elevated ethical standards emerge as distinct and noteworthy considerations, setting Healthcare apart from AI applications in other domains and more.

**Table 7**
Healthcare data can be used to train LLMs.

| Data | Type | Size | Link |
|---|---|---|---|
| MIMIC-III | EHR | 58,976 hospital admissions | Link |
| MIMIC-IV | EHR | 11 years of hospital admissions | Link |
| CPRD [154] | EHR | over 2000 primary care practices | Link |
| PubMed | SL | 35M biomedical literature | Link |
| PMC | SL | 8 million articles | Link |
| RCT [155] | SL | 4528 abstract | Link |
| MS^2 [156] | SL | 470,402 abstract | Link |
| CDSR [157] | SL | 7805 abstract | Link |
| SumPubMed [158] | SL | 33,772 abstract | Link |
| The Pile | SL | 825 GB English text | Link |
| S2ORC [159] | SL | 63,709 abstract | Link |
| CORD-19 [160] | SL | 1M papers | Link |
| MeQSum [161] | MS | 1000 instances | Link |
| CHQ-Sum [162] | MS | 1507 instances | Link |
| UMLS | KB | 2M entities for 900K concepts | Link |
| MedDialog [163] | Dial. | 3.66 million conversations | Link |
| CovidDialog [164] | Dial. | 603 consultations | Link |
| Flashcards [117] | Dial. | 33 955 instances | Link |
| Wikidoc [117] | Dial. | 67 704 instances | Link |
| Wikidoc PI [117] | Dial. | 5942 instances | Link |
| MEDIQA [165] | Dial. | 2208 instances | Link |
| CORD-19 [160] | Dial. | 1 056 660 instances | Link |
| MMMLU [160] | Dial. | 3787 instances | Link |
| Pubmed Causal [166] | Dial. | 2446 instances | Link |
| ChatDoctor [167] | Dial. | 215 000 instances | Link |
| Alpaca-EN-AN [168] | Inst. | 52K instructions | Link |
| Alpaca-CH-AN [168] | Inst. | 52K instructions | Link |
| ShareGPT | Dial. | 61 653 long conversations | Link |
| COMETA [169] | Web | 800K Reddit posts | Link |
| WebText | Web | 40 GB of text | Link |
| OpenWebText | Web | 38 GB of text | Link |
| Colossal Corpus | Web | 806 GB of text | Link |
| OpenI | EHR | 3.7 million images | Link |
| U-Xray [170] | MM | 3955 reports and 7470 images | Link |
| ROCO [171] | MM | 81,000 radiology images and captions | Link |
| MedICaT [172] | MM | 17,000 images includes captions | Link |
| PMC-OA [173] | MM | 1.6M image-caption pairs | Link |
| CheXpert [174] | MM | 224,316 chest radiographs with reports | Link |
| PadChest [175] | MM | 160,000 images with related text | Link |
| MIMIC-CXR | MM | 227,835 imaging for 64,588 patients | Link |
| PMC-15M [176] | MM | 15 million Figure-caption pairs | Link |
| OpenPath [177] | MM | 208,414 pathology images and text | Link |
| Medtrinity [178] | MM | 25 million images and text | Link |
| MedPix 2.0 [179] | MM | 12,000 patient case scenarios | Link |
| MultiMed [180] | MM | 2.56 million samples with 10 modalities | Link |
| WorldMedQA-V [181] | MM | 568 QAs with medical images | Link |

☆ Although there are datasets available for Instruction Fine-Tuning, such as Multi-MedQA and the USMLE test, we have opted not to include them in this list. These datasets are typically employed for evaluation purposes rather than serving as primary resources for training. SL, MS, MM and KB means Scientific Literature, Medical Question Summarization, Multimodal, and Knowledge Base, respectively. Dial. and Inst. mean Dialogue and Instruction.

**Table 8**
The statistics of computation cost for existing Healthcare LLM.

| Model Name | Total data size | GPU type | GPU no. | GPU time |
|---|---|---|---|---|
| Visual Med-Alpaca | 54k data points | A100-80G | 4 | 2.51 h |
| GatorTron | >90 billion words | A100 | 992 | 6 days |
| Galactica | – | A100-80G | 128 | – |
| ChatDoctor | 100k conversations | A100 | 6 | 3 h |
| DoctorGLM | 3.5G | A100-80G | 1 | 8 h |
| PMC-LLaMA | 75B tokens | A100 | 8 | 7 days |
| Visual Med-Alpaca | 44.8MB* (without images) | A100-80G | 4 | 2.51 h |
| BianQue 1.0 | 9 million samples | RTX 4090 | 8 | 16 days |
| GatorTronGPT | 277B tokens | A100-80G | 560 | 26 days |
| HuatuoGPT | 226,042 instances | A100 | 8 | – |
| LLaVA-Med | 15M image-caption pairs | A100 | 8 | 15 h |
| Med-Flamingo | 1.3M image-caption pairs | A100-80G | 8 | 6.75 days |

## 5.1. Fairness

Fairness within the context of LLMs refers to the principle of equitably treating all users and preventing any form of unjust discrimination. This essential concept revolves around the mitigation of biases, aiming to guarantee that the outcomes produced by an AI system do not provide undue advantages or disadvantages to specific individuals or groups. These determinations should not be influenced by factors such as race, gender, socioeconomic status, or any other related attributes, e.g., different input languages and processing tasks, striving for an impartial and balanced treatment of all users. This fundamental tenet aligns with the broader objective of promoting equality and inclusivity for Healthcare LLMs.

The biases from LLMs can be attributed to the uneven distribution of demographic attributes in pre-training corpora. Such an argument also holds for the Healthcare sector [187]. As an example, neural models trained on publicly accessible chest X-ray datasets tend to exhibit underdiagnosis tendencies in marginalized communities, including female patients, Black patients, Hispanic patients, and those covered by Medicaid insurance [188]. These specific patient groups often experience systemic underrepresentation within the datasets, resulting in biased algorithms that may be susceptible to shifts in population demographics and disease prevalence. Furthermore, several global disease classification systems display limited intra-observer consensus, implying that an algorithm trained and assessed in one country may undergo evaluation under a dissimilar labeling framework in another country [189].

Current common practices to improve AI fairness in the Healthcare domain focus on pre-processing, in-processing, and post-processing [187]. Importance weighting is a pre-processing technique, which involves adjusting the significance of less frequent samples from protected subgroups. Similarly, resampling endeavors to rectify sample-selection bias by acquiring more equitable subsets of the initial training dataset and can be naturally employed to address the underrepresentation of specific subgroups.

For LLMs, bias mitigation methods are frequently studied in the context of instruction fine-tuning and prompt engineering. The representative technique for instruction fine-tuning is RLHF. In the case of InstructGPT, GPT-3 is refined through a process involving RLHF, specifically aimed at adhering to human instructions. The procedure involves three sequential steps: firstly, gathering human-authored demonstration data to guide GPT-3's learning; secondly, assembling comparative data consisting of model-generated outputs assessed by annotators to construct a reward model that predicts outputs preferred by humans; and lastly, fine-tuning policies based on this reward model. The aforementioned process offers a valuable chance to rebalance the data and incorporate additional security measures to prevent biased behavior in the model. However, it is important to note that obtaining demographic information can sometimes be challenging due to privacy and ethical concerns in medical practices. This creates an obstacle when we aim to ensure fairness while also protecting privacy.

## 5.2. Accountability

LLMs are prone to amplifying the inherent social biases present in their training data, and they may produce hallucinatory or counterfactual outputs. This issue is compounded by their lack of robustness, making them vulnerable to perturbations and deviations from expected performance, especially when faced with diverse inputs or scenarios. In the healthcare sector, these problems can have grave implications because the outputs of LLMs can directly impact people's health and even their lives. Consequently, ensuring accountability becomes a crucial concern when deploying LLMs in healthcare settings.

Effective accountability acts as a vital safeguard, ensuring that LLMs can be reliably integrated into the Healthcare field. Specifically, accountability entails that when healthcare LLMs err or yield undesirable outcomes, clear attribution of responsibility enables swift identification of the responsible parties. This facilitates prompt remedial actions and appropriate compensation for affected patients. Addressing these issues not only resolves specific problems but also helps prevent similar issues in the future, thereby enhancing both patient and public trust in healthcare LLM applications.

The hallucinations problem presents a main obstacle to accountable AI. In the evaluation conducted by the study [190], ChatGPT was evaluated using fact-based question-answering datasets, revealing that its performance did not exhibit enhancements in comparison to earlier versions. Consequently, the reliability of ChatGPT in tasks necessitating faithfulness is called into question. For instance, its potential fabrication of references in the context of scientific article composition [191] and the invention of fictitious legal cases within the legal domain [192] accentuate the potential risks associated with its use in critical domains.

Further, McKenna et al. [193] and Li et al. [194] investigate the root reason of hallucinations. These studies pinpoint the root cause of the hallucination problem: LLMs tend to memorize training data, especially in relation to word frequencies. This fundamental cause indicates that completely resolving the hallucination issue is challenging. Consequently, even the most advanced LLMs may still produce incorrect information. For such reason, we have to make an effective accountability before applying Healthcare LLMs in real medical scenarios.

Actually, accountability in AI is not just about correcting errors but also about implementing preventative measures that maintain trust and safety, particularly when AI decisions impact human lives. A direct preventive measure is to facilitate user participation in modeling decisions. The study [195] contended that enabling users to access human-generated source references is crucial for enhancing the reliability of the model's responses. The study [196] advocated for the involvement of both AI developers and system safety engineers in evaluating the moral accountability concerning patient harm. Additionally, they recommend a transition from a static assurance model to a dynamic one, recognizing that ensuring safety is an ongoing process and cannot be entirely resolved during the initial design phase of the AI system before its deployment.

The study [197] proposed a solution to tackle the issue of accountability, advocating for the education and training of prospective AI users to discern the appropriateness of relying on AI recommendations. However, imparting this knowledge to practitioners demands a considerable investment of effort. Healthcare professionals frequently grapple with overwhelming workloads and burnout, making comprehensive training on AI a significant challenge. Moreover, not all Healthcare practitioners possess adequate statistical training to comprehend the underlying mechanics of AI algorithms. In addition to education, the study [197] recommended the establishment of policies and mechanisms to ensure the protection of both clinicians and AI within the Healthcare domain.

*5.3. Transparency*

The limited transparency of neural networks has been widely criticized, presenting significant obstacles to their application in the Healthcare domain. LLMs and PLMs are complex neural network models, which further exacerbate the challenges associated with interpretability. In recent years, there have been efforts to understand the inner workings of PLMs in Healthcare contexts. Probing PLMs have been extensively employed to uncover the underlying factors contributing to their performance. For example, the study [198] examined PLMs' disease knowledge, while the study [199] conducted in-depth analyses of attention in protein Transformer models, yielding valuable insights into their mechanisms. In the general meaning learning domain, a transparent model is typically characterized by decision-making processes akin to those of white-box models, e.g., decision tree-based models or linear regression models. It often encompasses post hoc explanations [200], model-specific explanations [201] or model-agnostic explanations [202]. Sometimes, the explanation insights are derived from feature maps [203], generated natural language [204], factual and counterfactual examples [205], or decision-making evidence [206].

For PLMs, the study [200] introduced an innovative method accompanied by quantitative metrics aimed at mitigating the limitations observed in existing post hoc explanation approaches. These drawbacks include reliance on human judgment, the necessity for retraining, and issues related to data distribution shifts. The method allows for a quantitative assessment of interpretability methods without the need for retraining and effectively addresses distribution shifts between training and evaluation sets. In the era of LLMs, CoT prompting [6] has emerged as a potential method for providing a certain level of interpretability by generating reasoning steps. The technique empowers LLMs to break down complex, multi-step problems into more manageable intermediate steps. Moreover, it offers a transparent view of the LLM's behavior, shedding light on its potential process of arriving at a specific answer and offering insights for identifying and rectifying errors in the reasoning path. However, this approach faces two primary challenges: the high cost of annotations required for CoT and the evaluation of interpretability. Acquiring demonstrations with annotated reasoning steps is an expensive task, particularly in professional fields such as Healthcare. Additionally, evaluating the generated reasoning results as explainable justifications and ensuring their usability pose significant challenges.

*5.4. Ethics*

The ethical concerns about using LLMs for Healthcare have been widely discussed. Healthcare LLMs typically possess a wide range of patient characteristics, including clinical measurements, molecular signatures, demographic information, and even behavioral and sensory tracking data. It is crucial to acknowledge that these models are susceptible to memorize training data and simply reproducing it for users, resulting compromising the privacy of users.

As mentioned in Section 4.2, EHRs serve as important training data, alongside public scientific literature and web data. However, it is worth noting that EHRs may contain sensitive information such as patient visits and medical history, and exposing such data could lead to physical and mental harm to patients. It is important to recognize that de-identification techniques employed in EHR may not always guarantee complete safety. Recent studies have shown that there can be instances of data leakage from PLMs, allowing for the recovery of personal health information from models trained on such data sources [207]. Additionally, approaches such as KART [208] have been proposed to assess the vulnerability of sensitive information in biomedical PLMs using various attack strategies.

Medical applications inherently involve sensitive data privacy concerns that surpass other NLP tasks. Consequently, safeguarding privacy during the evaluation process becomes more important. One potential solution to address this challenge is the adoption of Federated Learning (FL) [209], which enable the implementation of large-scale evaluation systems while preserving privacy. By allowing the model to be trained directly on the devices where the data originates, FL keeps sensitive patient information localized, reducing the risk of data breaches. Moreover, it can help in creating more generalized and unbiased models by learning from a diverse array of decentralized data sources, thus covering a broader spectrum of patient conditions.

Summarily, it is imperative for stakeholders to engage in ethical reviews and updates of the guidelines governing the use of LLMs. This includes regular assessments of the models for biases, implementing rigorous privacy safeguards, and ensuring transparent and explainable AI systems. Moreover, active collaboration between ethicists, technologists, clinicians, and patients is necessary to harness the benefits of healthcare LLMs while minimizing their risks.

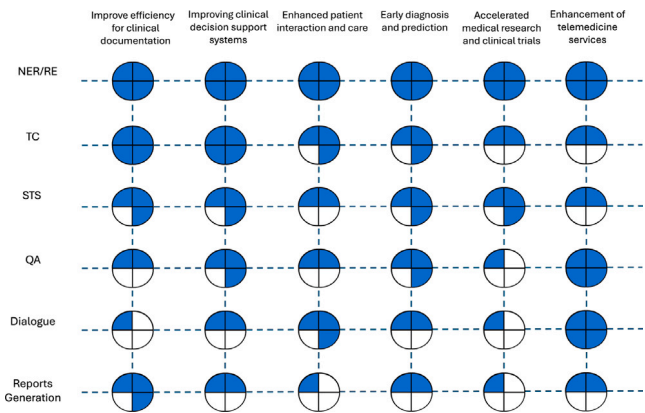

**Fig. 6.** The illustration depicts NLP technologies and their related healthcare applications. A quarter circle indicates that the technology is just beginning to be explored in these applications. Two quarters signify that the technology has been studied for several years. Three quarters suggest that the technology is mature and ready for implementation in real-world scenarios. A full circle indicates that the technology is actively being utilized in real scenarios.

## 6. Discussion

### 6.1. Healthcare core issues

As illustrated in Fig. 6, we identify six core issues in healthcare that are critical for improving healthcare outcomes. We then discuss how these core issues are supported by various LLM-related technologies introduced in Section 2. Generally, foundational technologies such as NER, RE, and TC are widely used in real-world scenarios. Furthermore, the generative capabilities of QA and dialogue systems play increasingly important roles in enhancing healthcare outcomes. The creation and management of clinical documentation are time-consuming, leading to inefficiencies and increased error risks. NER and RE automate the extraction of key information from medical notes, allowing medical professionals to focus more on patient care while reducing paperwork burdens. Also, LLMs can generate structured medical reports and ensure compliance with regulatory requirements, ultimately enhancing the quality of services and optimizing the healthcare system.

Clinical Decision Support Systems (CDSS) are vital in healthcare, assisting physicians with precise medical decisions through timely data analysis. Basic CDSS utilize NER and RE to extract key patient features, while STS analyzes similar patients for predictive outcomes. Advanced CDSS leverage LLMs for flexible decision support by addressing user-posed health queries, significantly enhancing the medical decision-making process despite their current rarity in practice. As the demand for healthcare services grows, traditional patient–doctor interaction face challenges, particularly for continuous care outside working hours. QA and Dialogue enable the creation of virtual health assistants that provide round-the-clock health consultations and medication management. These AI assistants can address common health issues, such as drug interactions and appointment management, although advanced features like emotional support are still under exploration.

Early diagnosis is vital for improving treatment outcomes, particularly for diseases like cancer, cardiovascular conditions, and brain problems [210–212]. By utilizing LLMs to analyze extensive historical health data, we can identify early disease signals and predict individual risks, employing NER/RE technologies to process structured data and TC for unstructured medical records, while QA and dialogue enhance accuracy in disease prediction.

In medical research, LLMs streamline the analysis of vast literature, using NER/RE to identify keywords and STS to find similar studies, significantly reducing literature review time. This acceleration facilitates access to the latest research findings, allowing researchers to directly query LLMs for multiple potential answers, which inspires further exploration and advancement in the field. The uneven distribution of healthcare resources limits access to medical services in remote areas, making it difficult for patients to receive timely care. QA and dialogue technologies enable chatbots to address common issues and recommend human experts for complex cases.

### 6.2. Multimodal healthcare LLMs

The healthcare domain inherently involves diverse multimodal data, making multimodal Healthcare LLMs one of the most promising and essential research directions [213]. By integrating textual data with medical images, time-series data, and other modalities, these models have the potential to deliver more comprehensive and insightful analyses.

On one hand, multimodal Healthcare LLMs, which can integrate and learn from heterogeneous data, offer the potential to unlock a profound and nuanced understanding of complex medical phenomena. By capturing complementary semantic information and the intricate relationships across various modalities, these models enable clinicians to gain a holistic view of patients' conditions. This capability supports more proactive monitoring, precise diagnoses, and highly personalized treatment plans. On the other hand, multimodal learning significantly broadens the application scope in the healthcare field. For instance, a patient's abdomen may develop a hard, lump-like protrusion, which ordinary patients might find difficult to describe accurately. In such cases, if an LLM could directly analyze the patient's photo to make a determination, its overall efficiency, capability, and practicality would be significantly enhanced.

Nevertheless, challenges such as data heterogeneity, integration complexity, and the need for large-scale, high-quality datasets persist [214]. Overcoming these challenges through continued research and innovation is vital to fully harness the transformative potential of multimodal data in healthcare LLMs.

#### 6.2.1. Integration with healthcare process

Is the application of artificial intelligence in the medical field just an "old myth", or can it really change the status quo? Clearly, although current AI solutions are fragmented and mostly experimental without widespread adoption, there exist such problems because we believe they are mainly caused by the following three reasons based on the existing study [215]. First, it is difficult to integrate with existing hospital information technology (IT) systems. AI solutions require large amounts of data for training, and most of this data is currently stored in hospitals' own information systems. Retrieving and integrating this data requires upgrades and modifications to existing systems, which will have an impact on hospitals' daily operations. In addition, different hospitals use different data formats and standards, lack standardized interfaces, and have relatively complex workflows in the Healthcare domain. AI systems find it difficult to adapt to different interfaces, which also increases the difficulty of integration. Second, fragmentation of IT systems due to hospital consolidations. With the increase in hospital mergers and acquisitions, the original hospitals may use completely different IT systems. After consolidation, it is necessary to unify their respective clinical and management systems, which requires huge investment and a long transition period. Introducing new AI systems during this process will face great technical challenges. Third, regulations are unclear and challenging. Currently, laws and regulations for AI medical applications are incomplete. Key issues such as information security, privacy protection, and liability attribution lack clear provisions. In addition, regulations differ across countries and regions. These will bring uncertainties to the development and application of AI systems. At the same time, the application of AI in the medical industry involves complex ethical issues that are also difficult to resolve.

*6.3. Global collaboration and regulatory differences*

In addition to general concerns about fairness, accountability, transparency, and ethics, differences between countries pose significant challenges to applying Healthcare LLMs, particularly in the context of global collaborations. One major barrier is the disparity in levels of digital development, often referred to as the "digital divide" [216]. Bridging this divide requires strategies to make LLM technologies more accessible and equitable, especially in under-resourced settings. This can be achieved by developing user-friendly interfaces, supporting multiple languages, and training models on diverse datasets that reflect the needs of various populations. Such inclusivity enhances the relevance and applicability of LLMs in global healthcare contexts.

Another critical challenge is addressing differences in global regulatory frameworks. Adapting LLMs to comply with diverse legal and ethical standards across regions requires a comprehensive understanding of each jurisdiction's regulations and cultural nuances. This adaptation not only ensures compliance with local legal frameworks but also fosters trust by respecting regional ethical considerations. Cross-national collaboration is pivotal in overcoming these challenges. Establishing shared governance models and standardized protocols can facilitate the seamless integration of LLMs across borders. Additionally, leveraging privacy-preserving technologies, such as federated learning, enables secure data sharing and collaborative model training while safeguarding patient confidentiality. These collaborative efforts can drive the development of robust, globally applicable healthcare solutions that are sensitive to regional differences and capable of addressing disparities in healthcare access and quality.

## 7. Conclusion

In this study, we provided a comprehensive survey specifically focusing on Healthcare LLMs. Our survey encompassed an extensive examination of data, technologies, applications, fairness, accountability, transparency, and ethics associated with Healthcare LLMs. A noteworthy transformation has been observed from Discriminative AI to Generative AI, as well as from model-centered to data-centered approaches, marking a significant shift from PLMs to LLMs. This transition has enabled Healthcare LLMs to support more advanced applications beyond conventional NLP-based fundamental tasks.

However, despite the opportunities presented by Healthcare LLMs, several significant challenges persist. Issues pertaining to interpretability, privacy protection, medical knowledge enhancement, integration with Healthcare processes, and effective interaction with patients and doctors pose substantial obstacles. These challenges hinder the translation of innovative LLMs into practical adoption within the Healthcare field. Consequently, physicians and other Healthcare professionals must carefully consider the potential benefits and limitations associated with LLMs as they navigate the selection and integration of these models into their medical practice.

## CRediT authorship contribution statement

**Kai He:** Writing – review & editing, Writing – original draft, Investigation, Formal analysis, Conceptualization. **Rui Mao:** Writing – review & editing, Writing – original draft, Methodology, Conceptualization. **Qika Lin:** Writing – review & editing, Writing – original draft, Methodology, Conceptualization. **Yucheng Ruan:** Writing – review & editing, Writing – original draft. **Xiang Lan:** Writing – review & editing, Writing – original draft, Conceptualization. **Mengling Feng:** Writing – review & editing, Supervision, Funding acquisition, Conceptualization. **Erik Cambria:** Writing – review & editing, Supervision, Conceptualization.

## Declaration of competing interest

The authors declare that they have no known competing financial interests or personal relationships that could have appeared to influence the work reported in this paper.

## Acknowledgments

This work has been supported by the National Research Foundation Singapore under AI Singapore Programme (Award Number: AISG-GC-2019-001-2A and AISG2-TC-2022-004); The RIE2025 Industry Alignment Fund (I2101E0002 – Cisco-NUS Accelerated Digital Economy Corporate Laboratory).

## Data availability

No data was used for the research described in the article.

## References

[1] Karan Singhal, et al., Towards expert-level medical question answering with large language models, 2023, arXiv preprint arXiv:2305.09617.
[2] Jacob Devlin Ming-Wei Chang Kenton, Lee Kristina Toutanova, BERT: Pre-training of deep bidirectional transformers for language understanding, in: Proceedings of NAACL-HLT, 2019, pp. 4171–4186.
[3] Yinhan Liu, et al., RoBERTa: A robustly optimized BERT pretraining approach, 2019.
[4] Kai He, et al., Understanding the patient perspective of epilepsy treatment through text mining of online patient support groups, Epilepsy Behav. 94 (2019) 65–71.
[5] Tom Brown, et al., Language models are few-shot learners, Adv. Neural Inf. Process. Syst. 33 (2020) 1877–1901.
[6] Jason Wei, et al., Chain-of-thought prompting elicits reasoning in large language models, Adv. Neural Inf. Process. Syst. 35 (2022) 24824–24837.
[7] Hongjian Zhou, et al., A survey of large language models in medicine: Progress, application, and challenge, 2023, arXiv preprint arXiv:2311.05112.
[8] Michael Moor, et al., Med-flamingo: a multimodal medical few-shot learner, 2023, arXiv preprint arXiv:2307.15189.
[9] Chunyuan Li, et al., Llava-med: Training a large language-and-vision assistant for biomedicine in one day, 2023, arXiv preprint arXiv:2306.00890.
[10] Juexiao Zhou, et al., Path to medical AGI: Unify domain-specific medical LLMs with the lowest cost, 2023, arXiv preprint arXiv:2306.10765.
[11] Hongbo Zhang, et al., HuatuoGPT, towards taming language model to be a doctor, 2023, arXiv preprint arXiv:2305.15075.
[12] Weihao Gao, et al., OphGLM: Training an ophthalmology large language-and-vision assistant based on instructions and dialogue, 2023, arXiv preprint arXiv:2306.12174.
[13] Chen Yirong, et al., SoulChat: The "empathy" ability of the large model is improved by mixing and fine-tuning the data set of long text consultation instructions and multiple rounds of empathy dialogue, 2023.
[14] Jesutofunmi A. Omiye, et al., Large language models in medicine: the potentials and pitfalls, 2023.
[15] Linmei Hu, et al., A survey of knowledge enhanced pre-trained language models, IEEE Trans. Knowl. Data Eng. (2023).
[16] Shubo Tian, Qiao Jin, Lana Yeganova, Po-Ting Lai, Qingqing Zhu, Xiuying Chen, Yifan Yang, Qingyu Chen, Won Kim, Donald C. Comeau, et al., Opportunities and challenges for ChatGPT and large language models in biomedicine and health, Brief. Bioinform. 25 (1) (2024) bbad493.
[17] Wayne Xin Zhao, et al., A survey of large language models, 2023, arXiv preprint arXiv:2303.18223.
[18] Bonan Min, et al., Recent advances in natural language processing via large pre-trained language models: A survey, ACM Comput. Surv. (2021).
[19] Suhana Bedi, Yutong Liu, Lucy Orr-Ewing, Dev Dash, Sanmi Koyejo, Alison Callahan, Jason A. Fries, Michael Wornow, Akshay Swaminathan, Lisa Soleymani Lehmann, et al., Testing and evaluation of health care applications of large language models: a systematic review, JAMA (2024).
[20] Elizabeth C. Stade, Shannon Wiltsey Stirman, Lyle H. Ungar, Cody L. Boland, H. Andrew Schwartz, David B. Yaden, João Sedoc, Robert J. DeRubeis, Robb Willer, Johannes C. Eichstaedt, Large language models could change the future of behavioral healthcare: a proposal for responsible development and evaluation, NPJ Ment. Heal. Res. 3 (1) (2024) 12.
[21] P.M. Lavanya, E. Sasikala, Deep learning techniques on text classification using natural language processing (NLP) in social healthcare network: A comprehensive survey, in: 2021 3rd International Conference on Signal Processing and Communication, ICPSC, IEEE, 2021, pp. 603–609.
[22] Kai He, Lixia Yao, Knowledge enhanced coreference resolution via gated attention, in: 2022 IEEE International Conference on Bioinformatics and Biomedicine, BIBM, IEEE, 2022, pp. 2287–2293.
[23] Kai He, et al., Construction of genealogical knowledge graphs from obituaries: Multitask neural network extraction system, J. Med. Internet Res. 23 (8) (2021) e25670.

[24] Longxiang Xiong, et al., How can entities improve the quality of medical dialogue generation? in: 2023 2nd International Conference on Big Data, Information and Computer Network, BDICN, IEEE, 2023, pp. 225–229.
[25] David S. Wishart, et al., DrugBank: a comprehensive resource for in silico drug discovery and exploration, Nucleic Acids Res. 34 (suppl_1) (2006) D668–D672.
[26] Alexander Dunn, et al., Structured information extraction from complex scientific text with fine-tuned large language models, 2022, arXiv preprint arXiv:2212.05238.
[27] Monica Agrawal, et al., Large language models are few-shot clinical information extractors, in: Proceedings of the 2022 Conference on Empirical Methods in Natural Language Processing, 2022, pp. 1998–2022.
[28] Long Ouyang, et al., Training language models to follow instructions with human feedback, Adv. Neural Inf. Process. Syst. 35 (2022) 27730–27744.
[29] Sunil Kumar Prabhakar, Dong-Ok Won, Medical text classification using hybrid deep learning models with multihead attention, Comput. Intell. Neurosci. 2021 (2021).
[30] Mohammed Ali Al-Garadi, et al., Text classification models for the automatic detection of nonmedical prescription medication use from social media, BMC Med. Inform. Decis. Mak. 21 (1) (2021) 1–13.
[31] Lossio-Ventura, et al., A comparison of chatgpt and fine-tuned open pre-trained transformers (opt) against widely used sentiment analysis tools: Sentiment analysis of covid-19 survey data, JMIR Ment. Heal. 11 (2024) e50150.
[32] Xiaofei Sun, et al., Text classification via large language models, 2023, arXiv preprint arXiv:2305.08377.
[33] Han Wang, Canwen Xu, Julian McAuley, Automatic multi-label prompting: Simple and interpretable few-shot classification, in: Proceedings of the 2022 Conference of the North American Chapter of the Association for Computational Linguistics: Human Language Technologies, 2022, pp. 5483–5492.
[34] Majid Rastegar-Mojarad, et al., BioCreative/OHNLP challenge 2018, in: ACM-BCB 2018 - Proceedings of the 2018 ACM International Conference on Bioinformatics, Computational Biology, and Health Informatics, in: ACM-BCB 2018 - Proceedings of the 2018 ACM International Conference on Bioinformatics, Computational Biology, and Health Informatics, Association for Computing Machinery, Inc, 2018, p. 575.
[35] Diwakar Mahajan, et al., Identification of semantically similar sentences in clinical notes: Iterative intermediate training using multi-task learning, JMIR Med. Inform. 8 (11) (2020) e22508.
[36] Susannah Fox, Maeve Duggan, Health online 2013, 2012.
[37] Qian Liu, et al., Semantic matching in machine reading comprehension: An empirical study, Inf. Process. Manage. 60 (2) (2023) 103145.
[38] Ankit Pal, et al., Medmcqa: A large-scale multi-subject multi-choice dataset for medical domain question answering, in: Conference on Health, Inference, and Learning, PMLR, 2022, pp. 248–260.
[39] Qiao Jin, et al., Pubmedqa: A dataset for biomedical research question answering, 2019, arXiv preprint arXiv:1909.06146.
[40] Dan Hendrycks, et al., Measuring massive multitask language understanding, 2020, arXiv preprint arXiv:2009.03300.
[41] Alaleh Hamidi, Kirk Roberts, Evaluation of AI chatbots for patient-specific EHR questions, 2023, arXiv preprint arXiv:2306.02549.
[42] Quan Guo, et al., A medical question answering system using large language models and knowledge graphs, Int. J. Intell. Syst. 37 (11) (2022) 8548–8564.
[43] Tuong Do, Binh X. Nguyen, Erman Tjiputra, Minh Tran, Quang D. Tran, Anh Nguyen, Multiple meta-model quantifying for medical visual question answering, in: Medical Image Computing and Computer Assisted Intervention–MICCAI 2021: 24th International Conference, Strasbourg, France, September 27–October 1, 2021, Proceedings, Part V 24, Springer, 2021, pp. 64–74.
[44] Jiawei Chen, Dingkang Yang, Yue Jiang, Yuxuan Lei, Lihua Zhang, MISS: A generative pre-training and fine-tuning approach for med-VQA, in: International Conference on Artificial Neural Networks, Springer, 2024, pp. 299–313.
[45] Shaoxiong Ji, et al., MentalBERT: Publicly available pretrained language models for mental healthcare, 2021.
[46] Reena L. Pande, et al., Leveraging remote behavioral health interventions to improve medical outcomes and reduce costs, Am. J. Manag. Care 21 (2) (2015) e141–e151.
[47] David Milward, Martin Beveridge, Ontology-based dialogue systems, in: Proc. 3rd Workshop on Knowledge and Reasoning in Practical Dialogue Systems, IJCAI03, 2003, pp. 9–18.
[48] Lin Xu, et al., End-to-end knowledge-routed relational dialogue system for automatic diagnosis, in: Proceedings of the AAAI Conference on Artificial Intelligence, 2019, pp. 7346–7353.
[49] Wei Qin, et al., Read, diagnose and chat: Towards explainable and interactive LLMs-augmented depression detection in social media, 2023, arXiv preprint arXiv:2305.05138.
[50] Li Yunxiang, et al., Chatdoctor: A medical chat model fine-tuned on llama model using medical domain knowledge, 2023, arXiv preprint arXiv:2303.14070.
[51] Baoyu Jing, Pengtao Xie, Eric Xing, On the automatic generation of medical imaging reports, in: Proceedings of the 56th Annual Meeting of the Association for Computational Linguistics (Volume 1: Long Papers), 2018, pp. 2577–2586.
[52] Yuan Xue, Tao Xu, L. Rodney Long, Zhiyun Xue, Sameer Antani, George R. Thoma, Xiaolei Huang, Multimodal recurrent model with attention for automated radiology report generation, in: Medical Image Computing and Computer Assisted Intervention–MICCAI 2018: 21st International Conference, Granada, Spain, September 16-20, 2018, Proceedings, Part I, Springer, 2018, pp. 457–466.
[53] Jun Chen, et al., VisualGPT: Data-efficient adaptation of pretrained language models for image captioning, in: Proceedings of the IEEE/CVF Conference on Computer Vision and Pattern Recognition, CVPR, 2022, pp. 18030–18040.
[54] Sheng Wang, et al., Chatcad: Interactive computer-aided diagnosis on medical image using large language models, 2023, arXiv preprint arXiv:2302.07257.
[55] Zhihong Chen, Yan Song, Tsung-Hui Chang, Xiang Wan, Generating radiology reports via memory-driven transformer, in: Proceedings of the 2020 Conference on Empirical Methods in Natural Language Processing, EMNLP, 2020, pp. 1439–1449.
[56] Aaron Nicolson, et al., Improving chest X-Ray report generation by leveraging warm-starting, 2022, arXiv preprint arXiv:2201.09405.
[57] Zihao Zhao, et al., ChatCAD+: Towards a universal and reliable interactive CAD using LLMs, 2023, arXiv preprint arXiv:2305.15964.
[58] Yikuan Li, Shishir Rao, José Roberto Ayala Solares, Abdelaali Hassaine, Rema Ramakrishnan, Dexter Canoy, Yajie Zhu, Kazem Rahimi, Gholamreza Salimi-Khorshidi, BEHRT: transformer for electronic health records, Sci. Rep. 10 (1) (2020) 7155.
[59] Hoo-Chang Shin, Yang Zhang, Evelina Bakhturina, Raul Puri, Mostofa Patwary, Mohammad Shoeybi, Raghav Mani, BioMegatron: Larger biomedical domain language model, 2020.
[60] Yu Gu, et al., Domain-specific language model pretraining for biomedical natural language processing, ACM Trans. Comput. Heal. (Health) 3 (1) (2021) 1–23.
[61] Ibrahim Burak Ozyurt, On the effectiveness of small, discriminatively pre-trained language representation models for biomedical text mining, in: Proceedings of the First Workshop on Scholarly Document Processing, 2020, pp. 104–112.
[62] Kamal raj Kanakarajan, et al., BioELECTRA:Pretrained biomedical text encoder using discriminators, in: Proceedings of the 20th Workshop on Biomedical Language Processing, Association for Computational Linguistics, Online, 2021, pp. 143–154.
[63] Wissam Antoun, et al., AraBERT: Transformer-based model for arabic language understanding, in: Proceedings of the 4th Workshop on Open-Source Arabic Corpora and Processing Tools, with a Shared Task on Offensive Language Detection, 2020, pp. 9–15.
[64] Keno K. Bressem, Lisa C. Adams, Robert A. Gaudin, Daniel Tröltzsch, Bernd Hamm, Marcus R. Makowski, Chan-Yong Schüle, Janis L. Vahldiek, Stefan M. Niehues, Highly accurate classification of chest radiographic reports using a deep learning natural language model pre-trained on 3.8 million text reports, Bioinformatics 36 (21) (2020) 5255–5261.
[65] Kai He, et al., Virtual prompt pre-training for prototype-based few-shot relation extraction, Expert Syst. Appl. 213 (2023) 118927.
[66] Hongyi Yuan, et al., BioBART: Pretraining and evaluation of a biomedical generative language model, in: BioNLP 2022@ ACL 2022, 2022, p. 97.
[67] Michihiro Yasunaga, et al., Linkbert: Pretraining language models with document links, 2022, arXiv preprint arXiv:2203.15827.
[68] Giacomo Miolo, Giulio Mantoan, Carlotta Orsenigo, Electramed: a new pre-trained language representation model for biomedical nlp, 2021, arXiv preprint arXiv:2104.09585.
[69] Zheng Yuan, et al., Improving biomedical pretrained language models with knowledge, 2021, arXiv preprint arXiv:2104.10344.
[70] Shreyas Sharma, Ron Daniel Jr., BioFLAIR: Pretrained pooled contextualized embeddings for biomedical sequence labeling tasks, 2019, arXiv preprint arXiv:1908.05760.
[71] Shoya Wada, Toshihiro Takeda, Shiro Manabe, Shozo Konishi, Jun Kamohara, Yasushi Matsumura, Pre-training technique to localize medical bert and enhance biomedical bert, 2020, arXiv preprint arXiv:2005.07202.
[72] Long N. Phan, et al., Scifive: a text-to-text transformer model for biomedical literature, 2021, arXiv preprint arXiv:2106.03598.
[73] Jinhyuk Lee, et al., BioBERT: a pre-trained biomedical language representation model for biomedical text mining, Bioinformatics 36 (4) (2020) 1234–1240.
[74] Usman Naseem, Matloob Khushi, Vinay Reddy, Sakthivel Rajendran, Imran Razzak, Jinman Kim, Bioalbert: A simple and effective pre-trained language model for biomedical named entity recognition, in: 2021 International Joint Conference on Neural Networks, IJCNN, IEEE, 2021, pp. 1–7.
[75] Nina Poerner, et al., Inexpensive domain adaptation of pretrained language models: Case studies on biomedical NER and covid-19 QA, 2020, arXiv preprint arXiv:2004.03354.
[76] Patrick Lewis, Myle Ott, Jingfei Du, Veselin Stoyanov, Pretrained language models for biomedical and clinical tasks: understanding and extending the state-of-the-art, in: Proceedings of the 3rd Clinical Natural Language Processing Workshop, 2020, pp. 146–157.
[77] Usman Naseem, Adam G. Dunn, Matloob Khushi, Jinman Kim, Benchmarking for biomedical natural language processing tasks with a domain specific albert, BMC Bioinformatics 23 (1) (2022) 1–15.

[78] Yifan Peng, Shankai Yan, Zhiyong Lu, Transfer learning in biomedical natural language processing: an evaluation of BERT and ELMo on ten benchmarking datasets, 2019, arXiv preprint arXiv:1906.05474.
[79] Kexin Huang, et al., Clinicalbert: Modeling clinical notes and predicting hospital readmission, 2019, arXiv preprint arXiv:1904.05342.
[80] Kexin Huang, Abhishek Singh, Sitong Chen, Edward Moseley, Chih-Ying Deng, Naomi George, Charolotta Lindvall, Clinical XLNet: Modeling sequential clinical notes and predicting prolonged mechanical ventilation, in: Proceedings of the 3rd Clinical Natural Language Processing Workshop, 2020, pp. 94–100.
[81] Zeljko Kraljevic, et al., MedGPT: Medical concept prediction from clinical narratives, 2021, arXiv preprint arXiv:2107.03134.
[82] Boran Hao, et al., Enhancing clinical bert embedding using a biomedical knowledge base, in: 28th International Conference on Computational Linguistics, COLING 2020, 2020.
[83] Xiaoman Zhang, Chaoyi Wu, Ya Zhang, Yanfeng Wang, Weidi Xie, Knowledge-enhanced visual-language pre-training on chest radiology images, 2023.
[84] Yoshimasa Kawazoe, et al., A clinical specific BERT developed with huge size of Japanese clinical narrative, MedRxiv (2020).
[85] Ningyu Zhang, et al., Conceptualized representation learning for chinese biomedical text mining, 2020, arXiv preprint arXiv:2008.10813.
[86] Yiwen Meng, William Speier, Michael K. Ong, Corey W. Arnold, Bidirectional representation learning from transformers using multimodal electronic health record data to predict depression, IEEE J. Biomed. Health Inf. 25 (8) (2021) 3121–3129.
[87] Laila Rasmy, Yang Xiang, Ziqian Xie, Cui Tao, Degui Zhi, Med-BERT: pretrained contextualized embeddings on large-scale structured electronic health records for disease prediction, NPJ Digit. Med. 4 (1) (2021) 86.
[88] Fangyu Liu, et al., Self-alignment pretraining for biomedical entity representations, 2020, arXiv preprint arXiv:2010.11784.
[89] Zheng Yuan, et al., CODER: Knowledge-infused cross-lingual medical term embedding for term normalization, J. Biomed. Inform. 126 (2022) 103983.
[90] Yen-Pin Chen, Yi-Ying Chen, Jr-Jiun Lin, Chien-Hua Huang, Feipei Lai, et al., Modified bidirectional encoder representations from transformers extractive summarization model for hospital information systems based on character-level tokens (AlphaBERT): development and performance evaluation, JMIR Med. Inform. 8 (4) (2020) e17787.
[91] Suchin Gururangan, Ana Marasović, Swabha Swayamdipta, Kyle Lo, Iz Beltagy, Doug Downey, Noah A. Smith, Don't stop pretraining: Adapt language models to domains and tasks, 2020, arXiv preprint arXiv:2004.10964.
[92] Xing Meng, et al., Self-supervised contextual language representation of radiology reports to improve the identification of communication urgency, AMIA Summits Transl. Sci. Proc. 2020 (2020) 413.
[93] Elisa Terumi Rubel Schneider, João Vitor Andrioli de Souza, Julien Knafou, Lucas Emanuel Silva e Oliveira, Jenny Copara, Yohan Bonescki Gumiel, Lucas Ferro Antunes de Oliveira, Emerson Cabrera Paraiso, Douglas Teodoro, Cláudia Maria Cabral Moro Barra, BioBERTpt - a Portuguese neural language model for clinical named entity recognition, in: Proceedings of the 3rd Clinical Natural Language Processing Workshop, Association for Computational Linguistics, Online, 2020, pp. 65–72.
[94] Xi Yang, Jiang Bian, William R. Hogan, Yonghui Wu, Clinical concept extraction using transformers, J. Am. Med. Inform. Assoc. 27 (12) (2020) 1935–1942.
[95] Junshu Wang, et al., Cloud-based intelligent self-diagnosis and department recommendation service using Chinese medical BERT, J. Cloud Comput. 10 (2021) 1–12.
[96] Guillermo López-García, et al., Transformers for clinical coding in spanish, IEEE Access 9 (2021) 72387–72397.
[97] Liliya Akhtyamova, Named entity recognition in Spanish biomedical literature: Short review and BERT model, in: 2020 26th Conference of Open Innovations Association, FRUCT, IEEE, 2020, pp. 1–7.
[98] Jenny Copara, et al., Contextualized French language models for biomedical named entity recognition, in: Actes de la 6e Conférence Conjointe Journées d'Études sur la Parole (JEP, 33e Édition), Traitement Automatique des Langues Naturelles (TALN, 27e Édition), Rencontre des Étudiants Chercheurs en Informatique pour le Traitement Automatique des Langues (RÉCITAL, 22e Édition). Atelier DÉfi Fouille de Textes, ATALA et AFCP, Nancy, France, 2020, pp. 36–48.
[99] Nada Boudjellal, Huaping Zhang, Asif Khan, Arshad Ahmad, Rashid Naseem, Jianyun Shang, Lin Dai, ABioNER: a BERT-based model for Arabic biomedical named-entity recognition, Complexity 2021 (2021) 1–6.
[100] Nasrin Taghizadeh, et al., SINA-BERT: a BERT-based model for Arabic biomedical named-entity recognition, 2021, arXiv preprint arXiv:2104.07613.
[101] Martin Müller, et al., Covid-twitter-bert: A natural language processing model to analyse covid-19 content on twitter, Front. Artif. Intell. 6 (2023) 1023281.
[102] Youngduck Choi, et al., Learning low-dimensional representations of medical concepts, AMIA Summits Transl. Sci. Proc. 2016 (2016) 41.
[103] David L. Wheeler, et al., Database resources of the national center for biotechnology information, Nucleic Acids Res. 36 (suppl_1) (2007) D13–D21.
[104] Shuang Liu, et al., Preliminary study on the knowledge graph construction of Chinese ancient history and culture, Information 11 (4) (2020) 186.
[105] Chen Li, et al., BioModels Database: An enhanced, curated and annotated resource for published quantitative kinetic models, BMC Syst. Biol. 4 (1) (2010) 1–14.
[106] David S. Wishart, et al., DrugBank 5.0: a major update to the DrugBank database for 2018, Nucleic Acids Res. 46 (D1) (2018) D1074–D1082.
[107] Haochun Wang, et al., HuaTuo: Tuning LLaMA model with Chinese medical knowledge, 2023.
[108] Xi Yang, et al., GatorTron: A large language model for clinical natural language processing, MedRxiv (2022).
[109] Cheng Peng, et al., A study of generative large language model for medical research and healthcare, 2023, arXiv preprint arXiv:2305.13523.
[110] Ross Taylor, et al., Galactica: A large language model for science, 2022.
[111] Qianqian Xie, et al., Me LLaMA: Foundation large language models for medical applications, 2024.
[112] Yang Tan, et al., Medchatzh: a better medical adviser learns from better instructions, 2023, arXiv preprint arXiv:2309.01114.
[113] Yanis Labrak, et al., BioMistral: A collection of open-source pretrained large language models for medical domains, 2024.
[114] Shu Chang, et al., Visual med-alpaca: A parameter-efficient biomedical LLM with visual capabilities, 2023.
[115] Xidong Wang, et al., Apollo: An lightweight multilingual medical LLM towards democratizing medical AI to 6B people, 2024.
[116] Mingchen Li, Jiatan Huang, Jeremy Yeung, Anne Blaes, Steven Johnson, Hongfang Liu, Hua Xu, Rui Zhang, CancerLLM: A large language model in cancer domain, 2024, arXiv preprint arXiv:2406.10459.
[117] Tianyu Han, et al., MedAlpaca–an open-source collection of medical conversational AI models and training data, 2023, arXiv preprint arXiv:2304.08247.
[118] Chen Yirong, et al., BianQue-1.0: Improving the "question" ability of medical chat model through finetuning with hybrid instructions and multi-turn doctor QA datasets, 2023.
[119] Honglin Xiong, et al., Doctorglm: Fine-tuning your chinese doctor is not a herculean task, 2023, arXiv preprint arXiv:2304.01097.
[120] Fujian Jia, et al., OncoGPT: A medical conversational model tailored with oncology domain expertise on a large language model meta-AI (LLaMA), 2024.
[121] Karan Singhal, et al., Large language models encode clinical knowledge, 2022, arXiv preprint arXiv:2212.13138.
[122] Chaoyi Wu, et al., Pmc-llama: Further finetuning llama on medical papers, 2023, arXiv preprint arXiv:2304.14454.
[123] Sagar Goyal, et al., Healai: A healthcare llm for effective medical documentation, in: Proceedings of the 17th ACM International Conference on Web Search and Data Mining, 2024, pp. 1167–1168.
[124] Sara Pieri, et al., BiMediX: Bilingual medical mixture of experts LLM, 2024.
[125] Iker García-Ferrero, et al., Medical mT5: An open-source multilingual text-to-text LLM for the medical domain, 2024.
[126] Shihao Yang, et al., EpiSemoGPT: A fine-tuned large language model for epileptogenic zone localization based on seizure semiology with a performance comparable to epileptologists, MedRxiv (2024).
[127] Suhyeon Lee, et al., LLM-CXR: Instruction-finetuned LLM for CXR image understanding and generation, in: The Twelfth International Conference on Learning Representations, 2023.
[128] Junda Wang, et al., JMLR: Joint medical LLM and retrieval training for enhancing reasoning and professional question answering capability, 2024.
[129] Guangyu Wang, et al., ClinicalGPT: Large language models finetuned with diverse medical data and comprehensive evaluation, 2023, arXiv preprint arXiv:2306.09968.
[130] Subhabrata Mukherjee, et al., Polaris: A safety-focused LLM constellation architecture for healthcare, 2024.
[131] Songhua Yang, et al., Zhongjing: Enhancing the Chinese medical capabilities of large language model through expert feedback and real-world multi-turn dialogue, 2023.
[132] Qichen Ye, et al., Qilin-med: Multi-stage knowledge injection advanced medical large language model, 2024.
[133] Ashwin Kumar Gururajan, Enrique Lopez-Cuena, Jordi Bayarri-Planas, Adrian Tormos, Daniel Hinjos, Pablo Bernabeu-Perez, Anna Arias-Duart, Pablo Agustin Martin-Torres, Lucia Urcelay-Ganzabal, Marta Gonzalez-Mallo, et al., Aloe: A family of fine-tuned open healthcare LLMs, 2024, arXiv preprint arXiv:2405.01886.
[134] Arun James Thirunavukarasu, et al., Trialling a large language model (ChatGPT) in general practice with the Applied Knowledge Test: observational study demonstrating opportunities and limitations in primary care, JMIR Med. Educ. 9 (1) (2023) e46599.
[135] Aidan Gilson, et al., How does ChatGPT perform on the United States medical licensing examination? The implications of large language models for medical education and knowledge assessment, JMIR Med. Educ. 9 (1) (2023) e45312.
[136] Tiffany H. Kung, et al., Performance of ChatGPT on USMLE: Potential for AI-assisted medical education using large language models, PLoS Digit. Heal. 2 (2) (2023) e0000198.
[137] Douglas Johnson, et al., Assessing the accuracy and reliability of AI-generated medical responses: an evaluation of the Chat-GPT model, 2023.

[138] Jason Holmes, et al., Evaluating large language models on a highly-specialized topic, radiation oncology physics, 2023, arXiv preprint arXiv:2304.01938.
[139] Jamil S. Samaan, et al., Assessing the accuracy of responses by the language model ChatGPT to questions regarding bariatric surgery, Obes. Surg. (2023) 1–7.
[140] Dat Duong, et al., Analysis of large-language model versus human performance for genetics questions, Eur. J. Human Genet. (2023) 1–3.
[141] Joseph Chervenak, et al., The promise and peril of using a large language model to obtain clinical information: ChatGPT performs strongly as a fertility counseling tool with limitations, Fertil. Steril. (2023).
[142] Namkee Oh, et al., ChatGPT goes to the operating room: evaluating GPT-4 performance and its potential in surgical education and training in the era of large language models, Ann. Surg. Treat. Res. 104 (5) (2023) 269.
[143] Zhuo Wang, et al., Can LLMs like GPT-4 outperform traditional AI tools in dementia diagnosis? Maybe, but not today, 2023, arXiv preprint arXiv:2306.01499.
[144] Adi Lahat, et al., Evaluating the use of large language model in identifying top research questions in gastroenterology, Sci. Rep. 13 (1) (2023) 4164.
[145] Qing Lyu, et al., Translating radiology reports into plain language using chatgpt and gpt-4 with prompt learning: Promising results, limitations, and potential, 2023, arXiv preprint arXiv:2303.09038.
[146] Israt Jahan, et al., Evaluation of ChatGPT on biomedical tasks: A zero-shot comparison with fine-tuned generative transformers, 2023, arXiv preprint arXiv:2306.04504.
[147] Marco Cascella, et al., Evaluating the feasibility of ChatGPT in healthcare: an analysis of multiple clinical and research scenarios, J. Med. Syst. 47 (1) (2023) 33.
[148] Edward J. Hu, et al., Lora: Low-rank adaptation of large language models, 2021, arXiv preprint arXiv:2106.09685.
[149] Sewon Min, et al., Rethinking the role of demonstrations: What makes in-context learning work? 2022, arXiv preprint arXiv:2202.12837.
[150] Yoshua Bengio, et al., From system 1 deep learning to system 2 deep learning, in: Neural Information Processing Systems, 2019.
[151] Takeshi Kojima, et al., Large language models are zero-shot reasoners, 2023.
[152] Lilian Weng, LLM-powered autonomous agents, 2023, lilianweng.github.io.
[153] AutoGPT, AutoGPT official, 2024, https://autogpt.net/. (Accessed 12 September 2024).
[154] Emily Herrett, et al., Data resource profile: clinical practice research datalink (CPRD), Int. J. Epidemiol. 44 (3) (2015) 827–836.
[155] Byron C. Wallace, et al., Generating (factual?) narrative summaries of RCTs: Experiments with neural multi-document summarization, in: Proceedings of AMIA Informatics Summit, 2021.
[156] Jay DeYoung, et al., MS^2: Multi-document summarization of medical studies, in: Proceedings of the 2021 Conference on Empirical Methods in Natural Language Processing, Association for Computational Linguistics, Online and Punta Cana, Dominican Republic, 2021, pp. 7494–7513.
[157] Yue Guo, et al., Automated lay language summarization of biomedical scientific reviews, 2020, arXiv preprint arXiv:2012.12573.
[158] Vivek Gupta, et al., SUMPUBMED: Summarization dataset of PubMed scientific article, in: Proceedings of the 2021 Conference of the Association for Computational Linguistics: Student Research Workshop, Association for Computational Linguistics, 2021.
[159] Jennifer Bishop, et al., GenCompareSum: a hybrid unsupervised summarization method using salience, in: Proceedings of the 21st Workshop on Biomedical Language Processing, 2022, pp. 220–240.
[160] Lucy Lu Wang, et al., CORD-19: The COVID-19 open research dataset, in: Proceedings of the 1st Workshop on NLP for COVID-19 at ACL 2020, Association for Computational Linguistics, Online, 2020.
[161] Asma Ben Abacha, Dina Demner-Fushman, On the summarization of consumer health questions, in: Proceedings of the 57th Annual Meeting of the Association for Computational Linguistics, ACL 2019, Florence, Italy, July 28th - August 2, 2019.
[162] Shweta Yadav, et al., Chq-summ: A dataset for consumer healthcare question summarization, 2022, arXiv preprint arXiv:2206.06581.
[163] Guangtao Zeng, et al., MedDialog: Large-scale medical dialogue datasets, in: Proceedings of the 2020 Conference on Empirical Methods in Natural Language Processing, EMNLP, 2020, pp. 9241–9250.
[164] Zeqian Ju, et al., Coviddialog: Medical dialogue datasets about covid-19, 2020.
[165] Max Savery, et al., Question-driven summarization of answers to consumer health questions, Sci. Data 7 (1) (2020) 322.
[166] Bei Yu, et al., Detecting causal language use in science findings, in: Proceedings of the 2019 Conference on Empirical Methods in Natural Language Processing and the 9th International Joint Conference on Natural Language Processing, EMNLP-IJCNLP, Association for Computational Linguistics, Hong Kong, China, 2019, pp. 4664–4674.
[167] Yunxiang Li, et al., ChatDoctor: A medical chat model fine-tuned on a large language model meta-AI (LLaMA) using medical domain knowledge, Cureus 15 (6) (2023).
[168] Rohan Taori, et al., Stanford alpaca: An instruction-following LLaMA model, 2023, https://github.com/tatsu-lab/stanford_alpaca.
[169] Marco Basaldella, et al., COMETA: A corpus for medical entity linking in the social media, 2020, arXiv preprint arXiv:2010.03295.
[170] Dina Demner-Fushman, et al., Preparing a collection of radiology examinations for distribution and retrieval, J. Am. Med. Inform. Assoc. 23 (2) (2016) 304–310.
[171] Obioma Pelka, et al., Radiology Objects in COntext (ROCO): a multimodal image dataset, in: Intravascular Imaging and Computer Assisted Stenting and Large-Scale Annotation of Biomedical Data and Expert Label Synthesis: 7th Joint International Workshop, CVII-STENT 2018 and Third International Workshop, LABELS 2018, Held in Conjunction with MICCAI 2018, Granada, Spain, September 16, 2018, Proceedings 3, Springer, 2018, pp. 180–189.
[172] Sanjay Subramanian, et al., Medicat: A dataset of medical images, captions, and textual references, 2020, arXiv preprint arXiv:2010.06000.
[173] Weixiong Lin, et al., Pmc-clip: Contrastive language-image pre-training using biomedical documents, 2023, arXiv preprint arXiv:2303.07240.
[174] Jeremy Irvin, et al., Chexpert: A large chest radiograph dataset with uncertainty labels and expert comparison, in: Proceedings of the AAAI Conference on Artificial Intelligence, Vol. 33, 2019, pp. 590–597.
[175] Aurelia Bustos, et al., Padchest: A large chest x-ray image dataset with multi-label annotated reports, Med. Image Anal. 66 (2020) 101797.
[176] Sheng Zhang, et al., Large-scale domain-specific pretraining for biomedical vision-language processing, 2023, arXiv preprint arXiv:2303.00915.
[177] Zhi Huang, et al., A visual–language foundation model for pathology image analysis using medical Twitter, Nature Med. (2023).
[178] Yunfei Xie, Ce Zhou, Lang Gao, Juncheng Wu, Xianhang Li, Hong-Yu Zhou, Sheng Liu, Lei Xing, James Zou, Cihang Xie, et al., Medtrinity-25m: A large-scale multimodal dataset with multigranular annotations for medicine, 2024, arXiv preprint arXiv:2408.02900.
[179] Irene Siragusa, Salvatore Contino, Massimo La Ciura, Rosario Alicata, Roberto Pirrone, MedPix 2.0: A comprehensive multimodal biomedical dataset for advanced AI applications, 2024, arXiv preprint arXiv:2407.02994.
[180] Shentong Mo, Paul Pu Liang, MultiMed: Massively multimodal and multitask medical understanding, 2024, arXiv preprint arXiv:2408.12682.
[181] João Matos, Shan Chen, Siena Placino, Yingya Li, Juan Carlos Climent Pardo, Daphna Idan, Takeshi Tohyama, David Restrepo, Luis F. Nakayama, Jose M.M. Pascual-Leone, et al., WorldMedQA-V: a multilingual, multimodal medical examination dataset for multimodal language models evaluation, 2024, arXiv preprint arXiv:2410.12722.
[182] Kamal Raj Kanakarajan, et al., BioELECTRA: pretrained biomedical text encoder using discriminators, in: Proceedings of the 20th Workshop on Biomedical Language Processing, 2021, pp. 143–154.
[183] Marco Basaldella, et al., COMETA: A corpus for medical entity linking in the social media, in: Proceedings of the 2020 Conference on Empirical Methods in Natural Language Processing, EMNLP, Association for Computational Linguistics, Online, 2020, pp. 3122–3137.
[184] Tanmay Chavan, et al., A Twitter BERT approach for offensive language detection in Marathi, 2022, arXiv preprint arXiv:2212.10039.
[185] Xinyang Zhang, et al., TwHIN-BERT: A socially-enriched pre-trained language model for multilingual tweet representations, 2022, arXiv preprint arXiv:2209.07562.
[186] Claudia Wagner, et al., Measuring algorithmically infused societies, Nature 595 (7866) (2021) 197–204.
[187] Richard J. Chen, et al., Algorithmic fairness in artificial intelligence for medicine and healthcare, Nat. Biomed. Eng. 7 (6) (2023) 719–742.
[188] Laleh Seyyed-Kalantari, et al., Underdiagnosis bias of artificial intelligence algorithms applied to chest radiographs in under-served patient populations, Nature Med. 27 (12) (2021) 2176–2182.
[189] Alexandre Loupy, et al., Thirty years of the International Banff Classification for Allograft Pathology: the past, present, and future of kidney transplant diagnostics, Kidney Int. 101 (4) (2022) 678–691.
[190] Terry Yue Zhuo, et al., Red teaming ChatGPT via jailbreaking: Bias, robustness, reliability and toxicity, 2023, arXiv preprint arXiv:2301.12867.
[191] Sai Anirudh Athaluri, et al., Exploring the boundaries of reality: Investigating the phenomenon of artificial intelligence hallucination in scientific writing through ChatGPT references, Cureus 15 (4) (2023).
[192] Aniket Deroy, et al., How ready are pre-trained abstractive models and LLMs for legal case judgement summarization? 2023, arXiv preprint arXiv:2306.01248.
[193] Nick McKenna, et al., Sources of hallucination by large language models on inference tasks, 2023, arXiv preprint arXiv:2305.14552.
[194] Yifan Li, et al., Evaluating object hallucination in large vision-language models, 2023, arXiv preprint arXiv:2305.10355.
[195] Rui Mao, Guanyi Chen, Xulang Zhang, Frank Guerin, Erik Cambria, GPTEval: A survey on assessments of ChatGPT and GPT-4, in: Proceedings of the 2024 Joint International Conference on Computational Linguistics, Language Resources and Evaluation, LREC-COLING 2024, ELRA and ICCL, Torino, Italia, 2024, pp. 7844–7866.
[196] Ibrahim Habli, et al., Artificial intelligence in health care: accountability and safety, Bull. World Health Organ. 98 (4) (2020) 251.
[197] Avishek Choudhury, et al., Impact of accountability, training, and human factors on the use of artificial intelligence in healthcare: Exploring the perceptions of healthcare practitioners in the US, Hum. Factors Heal. 2 (2022) 100021.

[198] Liliya Akhtyamova, Paloma Martínez, Karin Verspoor, John Cardiff, Testing contextualized word embeddings to improve NER in Spanish clinical case narratives, IEEE Access 8 (2020) 164717–164726.
[199] Hao Tan, Mohit Bansal, Lxmert: Learning cross-modality encoder representations from transformers, 2019, arXiv preprint arXiv:1908.07490.
[200] Hugues Turbé, et al., Evaluation of post-hoc interpretability methods in time-series classification, Nat. Mach. Intell. 5 (3) (2023) 250–260.
[201] Sooji Han, et al., Hierarchical attention network for explainable depression detection on Twitter aided by metaphor concept mappings, in: Proceedings of the 29th International Conference on Computational Linguistics, COLING, International Committee on Computational Linguistics, Gyeongju, Republic of Korea, 2022, pp. 94–104.
[202] Marco Tulio Ribeiro, et al., "Why should I trust you?" Explaining the predictions of any classifier, in: Proceedings of the 22nd ACM SIGKDD International Conference on Knowledge Discovery and Data Mining, 2016, pp. 1135–1144.
[203] Rui Mao, et al., Word embedding and WordNet based metaphor identification and interpretation, in: Proceedings of the 56th Annual Meeting of the Association for Computational Linguistics, ACL, Vol. 1, Association for Computational Linguistics, Melbourne, Australia, 2018, pp. 1222–1231.
[204] Mengshi Ge, et al., Explainable metaphor identification inspired by conceptual metaphor theory, in: Proceedings of the AAAI Conference on Artificial Intelligence, Vol. 36, 2022, pp. 10681–10689, (10).
[205] Soo Hyun Cho, Kyung-shik Shin, Feature-weighted counterfactual-based explanation for bankruptcy prediction, Expert Syst. Appl. 216 (2023) 119390.
[206] Wei Li, et al., SKIER: A symbolic knowledge integrated model for conversational emotion recognition, in: Proceedings of the AAAI Conference on Artificial Intelligence, Vol. 37, 2023, pp. 13121–13129, (11).
[207] Jie Huang, Hanyin Shao, Kevin Chen-Chuan Chang, Are large pre-trained language models leaking your personal information? 2022.
[208] Yuta Nakamura, et al., KART: Parameterization of privacy leakage scenarios from pre-trained language models, 2020, arXiv preprint arXiv:2101.00036.
[209] Chen Zhang, et al., A survey on federated learning, Knowl.-Based Syst. 216 (2021) 106775.
[210] Jiaxing Xu, Kai He, Mengcheng Lan, Qingtian Bian, Wei Li, Tieying Li, Yiping Ke, Miao Qiao, Contrasformer: A brain network contrastive transformer for neurodegenerative condition identification, in: Proceedings of the 33rd ACM International Conference on Information and Knowledge Management, 2024, pp. 2671–2681.
[211] Zeyu Gao, Anyu Mao, Yuxing Dong, Jialun Wu, Jiashuai Liu, Chunbao Wang, Kai He, Tieliang Gong, Chen Li, Mireia Crispin-Ortuzar, Accurate spatial quantification in computational pathology with multiple instance learning, MedRxiv (2024).
[212] Jiaxing Xu, Mengcheng Lan, Xia Dong, Kai He, Wei Zhang, Qingtian Bian, Yiping Ke, Multi-atlas brain network classification through consistency distillation and complementary information fusion, 2024, arXiv preprint arXiv:2410.08228.
[213] Qika Lin, Yifan Zhu, Xin Mei, Ling Huang, Jingying Ma, Kai He, Zhen Peng, Erik Cambria, Mengling Feng, Has multimodal learning delivered universal intelligence in healthcare? A comprehensive survey, Inf. Fusion (2024) 102795.
[214] Jialun Wu, Xinyao Yu, Kai He, Zeyu Gao, Tieliang Gong, PROMISE: A pretrained knowledge-infused multimodal representation learning framework for medication recommendation, Inf. Process. Manage. 61 (4) (2024) 103758.
[215] Arya S. Rao, et al., Assessing the utility of ChatGPT throughout the entire clinical workflow, MedRxiv (2023).
[216] Sy Atezaz Saeed, Ross MacRae Masters, Disparities in health care and the digital divide, Curr. Psychiatry Rep. 23 (2021) 1–6.